\documentclass[final,5p,times,twocolumn,authoryear]{elsarticle}
\usepackage{amssymb}
\usepackage[caption=false,font=normalsize,labelfont=sf,textfont=sf]{subfig}
\usepackage{booktabs} 
\usepackage{multirow}
\usepackage{color}
\usepackage{amsmath,amsfonts}
\usepackage{cleveref}
\usepackage{ulem}
\usepackage{algorithmic}
\usepackage{algorithm}

\journal{Journal}

\begin{document}

\begin{frontmatter}
\title{A Benchmarking Protocol for SAR Colorization: From Regression to Deep Learning Approaches}

\author[1]{Kangqing Shen}

\author[2,3]{Gemine Vivone}

\author[1,4]{Xiaoyuan Yang \corref{cor1}}
\ead{<xiaoyuanyang@vip.163.com>}

\author[2,5]{Simone Lolli}

\author[6]{Michael Schmitt}

\affiliation[1]{addressline={School of Mathematical Sciences, Beihang University}, 
	city={Beijing},
	postcode={102206}, 
	country={China}}

\affiliation[2]{
	addressline={Institute of Methodologies for Environmental Analysis, CNR-IMAA}, 
	city={Tito Scalo},
	postcode={85050}, 
	country={Italy}}

\affiliation[3]{
	addressline={National Biodiversity Future Center, NBFC}, 
	city={Palermo},
	postcode={90133}, 
	country={Italy}}

\affiliation[4]{
	addressline={Key Laboratory of Mathematics, Information and Behavior, Ministry of Education, Beihang University}, 
	city={Beijing},
	postcode={102206}, 
	country={China}}

\affiliation[5]{
	addressline={CommSensLab, Department of Signal Theory and Communications, Polytechnic University of Catalonia}, 
	city={Barcelona},
	postcode={08034}, 
	country={Spain}}

\affiliation[6]{
	addressline={University of the Bundeswehr}, 
	city={Neubiberg},
	postcode={85577}, 
	country={Germany}}

\cortext[cor1]{Corresponding author.}


\begin{abstract}
Synthetic aperture radar (SAR) images are widely used in remote sensing. Interpreting SAR images can be challenging due to their intrinsic speckle noise and grayscale nature. To address this issue, SAR colorization has emerged as a research direction to colorize gray scale SAR images while preserving the original spatial information and radiometric information. However, this research field is still in its early stages, and many limitations can be highlighted. In this paper, we propose a full research line for supervised learning-based approaches to SAR colorization. Our approach includes a protocol for generating synthetic color SAR images, several baselines, and an effective method based on the conditional generative adversarial network (cGAN) for SAR colorization. We also propose numerical assessment metrics for the problem at hand. To our knowledge, this is the first attempt to propose a research line for SAR colorization that includes a protocol, a benchmark, and a complete performance evaluation. Our extensive tests demonstrate the effectiveness of our proposed cGAN-based network for SAR colorization. The code will be made publicly available.
\end{abstract}


\end{frontmatter}

\section{Introduction}
Thanks to the rapid advancements in remote sensing imaging methods, such as panchromatic, multispectral, hyperspectral, infrared, synthetic aperture radar (SAR), and night light imaging, researchers now have more rapid access to various sources of remote sensing data. However, the single different kinds of data only describe the observed scenes from a specific point of view, which may limit the applications; on the other hand, the multiple information sources provided by these images are redundant and complementary. Consequently, by integrating image information obtained from different sensors, fusion techniques can achieve more accurate and comprehensive remote sensing observations \citep{fusionreview_liu2022a,fusionreview_Zhang2021b,fusionreview_li2022a,fusionreview_zhu2017,vivone2023multispectral}.

Multi-source image fusion can be categorized into four types: $i$) homogeneous remote sensing data fusion; $ii$) heterogeneous remote sensing data fusion; $iii$) remote sensing site data fusion; and $iv$) remote sensing-non-observed data fusion. The goal of remote sensing research includes the fusion of homogeneous and heterogeneous data. Pansharpening is an example of homogeneous data fusion and has been studied extensively, being one of the first areas of research \citep{pansharpeningreview_vivone2014,pansharpeningreview_vivone2020,pansharpeningreview_DadrassJavan2021,pansharpeningreview_vivone2021}. On the contrary, SAR and optical image fusion is the fusion of heterogeneous remote sensing data, which is a challenging task due to the unique characteristics of SAR imagery. Recent studies focused on this research topic \citep{sarfusreview_kulkarni2020,sarfusreview_schmitt2017,sarfusreview_zhu2021}.

SAR image is an active microwave-based imaging. Being the wavelength longer, images are not affected by clouds, haze, and other meteorological conditions that otherwise affect visible images \citep{loll17}. The atmosphere is then transparent on SAR images except for heavy rain. Thanks to this feature, SAR images can be independently acquired day and night under almost all environmental conditions. SAR images mainly reveal the structural properties of the target scene, such as dielectric properties, surface roughness, and moisture content \citep{mace00}. Therefore, from SAR images it is possible to detect various objects based on their surface features containing rich spatial information. However, disadvantages of SAR images are not irrelevant; in fact, the lack of color information and severe speckle noise make image interpretation a challenging task even for well-trained remote sensing experts \citep{sardataset_schmitt2018}.

Unlike SAR images, passive optical satellite sensors record the electromagnetic spectrum reflected by the observed scene. Multispectral (MS) images are among the most important optical images due to their richness in spectral information. Therefore, by using the complementary characteristics of these two types of images, understanding and interpretation of SAR images would be much easier. SAR-MS image fusion can be a solution \citep{sarfus_kong2021,sarfus_ye2022, sarfus_chibani2006,sarfus_zhang2022}. However, it requires that the two source images be matched accurately and simultaneously acquired, i.e. general fusion methods cannot particularly help the interpretation of SAR images seen as an independent data source.

To overcome the dependency on paired SAR-MS images, SAR colorization is a promising technique that can be used to learn practical colorization for SAR images. As shown in \citep{cyclegan_sarcol_lee2021}, the colorization of the SAR can be divided into two categories. The first category is based on radar polarimetry theory. In \citep{traditional_sarcol_deng2008}, SAR images are colorized according to the fundamental principle that pixels that exhibit similar scattering properties are assigned to a similar hue. This way enables the reconstruction of complete polarimetric information from non-full polarimetric SAR images, but its applicability is strictly constrained to some experts in the field due to its technical complexity. The second category considers the task of SAR colorization as an image-to-image problem and employs neural networks to generate colorized SAR images.

To our knowledge, the literature on SAR colorization is limited. Inspired by \citep{divcolor_deshpande2017}, \citep{diverse_sarcol_schmitt2018} introduces the first deep learning-based SAR colorization methodology (DivColSAR), which utilizes a variational autoencoder (VAE) and a mixed density network (MDN) to generate multimodal colorization hypotheses. Since there are no ideal color SAR images available for ground truth during supervised training, they create a pseudo-SAR image using a Lab transform-based fusion algorithm. However, the proposed solution has several limitations. First, their article lacks quantitative evaluation. Furthermore, the visual results of the proposed method are not compared against those of any other technique.

In the current literature, there is often confusion about SAR colorization and SAR-to-optical image translation. Thus, we would like to first highlight the differences between the two terms, then reviewing instances into the two above-mentioned classes. Indeed, SAR-to-optical image translation main goal is to transform the SAR image into the correspondent optical image (i.e., even losing some spatial features (e.g., speckling) of a classical SAR image). This implies that the objective of the translation task is to generate an output image as similar as possible to the (reference) optical image. Instead, the aim of SAR colorization is to obtain a colorized SAR image that contains both SAR image information (including effects as shadowing, layover, foreshortening, and SAR noise as speckling) and color information coming from the related optical image (but not reproducing the optical image itself). Hence, we have that there are not available labels for SAR colorization training. Instead, labels are accessible for SAR-to-optical image translation.
Based on the analysis above, we can say that although \citep{mcgan_sarcol_ji2021,cyclegan_sarcol_lee2021,lee2022labeling,ku2018method} state that their own approaches are for SAR colorization, they are instead related to the SAR-to-optical image translation task. Their training procedure is devoted to the translation from SAR to optical images. Besides, the assessment process relies upon the comparison between the output image and the ground truth (optical) data. Finally, upon examining the outcomes, it becomes clear that all radar effects and speckle are wiped out during the translation between the SAR and optical domains. Thus, these approaches are improperly assigned to the SAR colorization class but, instead, they belong more to the SAR-to-optical image translation family. Thus, except for this work and the pioneering paper in \citep{diverse_sarcol_schmitt2018}, no other paper can be found in literature that deeply investigates the SAR colorization problem in the framework of the second category. Indeed, all the other machine learning-based works, claiming to perform SAR image colorization, just convert the SAR image into a pseudo-optical color image losing all SAR image special features.
Nevertheless, these misclassified SAR-to-optical image translation works are still valuable, which could be used to borrow some techniques for the SAR colorization task, such as CycleGAN architecture.

The main contributions of this work can be described as follows:
\begin{enumerate}[1)]
\item	We propose a full framework for SAR colorization supervised learning-based approaches including:
	\begin{itemize}
		\item A protocol to generate synthetic colorized SAR images.
		\item Several baselines from the simple linear regression to recent convolutional neural network architectures.
		\item Multidimensional similarity quality metrics to numerically assess (with reference) the quality of colorized SAR images.
	\end{itemize}
\item An effective cGAN-based method (adopted to address the specific SAR colorization problem) is also proposed, strongly outperforming the current state-of-the-art for SAR colorization using neural networks (i.e., DivColSAR \citep{diverse_sarcol_schmitt2018}).
\end{enumerate}

To the best of our knowledge, this represents the first endeavor to introduce a framework for SAR colorization that encompasses a protocol, a comprehensive benchmark, and a thorough performance assessment. This, in turn, lays the foundation for further research in this field.

The paper is organized as follows. In Section \ref{related work}, we introduce some related works. Section \ref{protocol} introduces the protocol for colorizing SAR images and the related performance metrics. Section \ref{baseline_method} describes the proposed solutions. Section \ref{result} presents the experimental results. Section \ref{discussion} is related to some discussions, while Section \ref{conclusion} draws the conclusions and future developments of this work.

\section{Related works} \label{related work}
Generative adversarial network (GAN) has emerged as a formidable class of machine learning models used for the synthesis of artificial data. The pioneering work of Goodfellow \citep{gan_goodfellow2014} introduces the GAN framework, which has found extensive applications in various image processing domains, including image super resolution \citep{esrgan_wang2018,realesrgan_wang2021}, pansharpening \citep{psgan_Liu2021a,pgman_Zhou2021,ucgan_zhou2022a, pancorlorgan_ozcelik2021}, and image-to-image translation \citep{img2imggan_chen2020,img2imggan_liu2017unsupervised,sar2optgan_wang2022a,sar2optgan_yang2022a}. GAN consists of two primary components, namely a generator that produces synthetic images that are perceptually indistinguishable from authentic ones and a discriminator that discriminates between real and synthetic data. However, the original GAN is plagued by unstable training issues and suboptimal control of the generated output.

Conditional generative adversarial network (cGAN) \citep{cgan_mirza2014} represents an extension of the GAN model that allows learning of data generation conditioned on specific input variables. This means that the generated samples can be tailored to a particular context. For example, in image generation, a label or a specific characteristic can be provided and the generator will produce an image that matches the condition. In particular, cGAN has an advantage over GAN because they can produce high-quality images with fewer iterations. This happens because the conditional input narrows the search space of the generator network, allowing a high-quality output to be achieved. For this reason, the conditional input facilitates the generation of high-quality outputs. This characteristic also contributes to the attainment of superior training stability. Assume that $G$ and $D$ represent the generator and discriminator, respectively. The loss function of cGAN can be formulated as
\begin{equation}
	\begin{aligned}
	\min _G \max _D \mathcal{L}_{c G A N}(G, D)= & E[\log (D(\mathbf{X}, \mathbf{Y}))] \\ 
	& +E[\log (1-D(\mathbf{X}, G(\mathbf{X})))],
	\end{aligned}
\end{equation}
where $\mathbf{Y}$ is the ground truth, $\mathbf{X}$ is the conditional input, $G(\mathbf{X})$ is the generated image of the generator, while $E$ and $\log$ represent the expectation symbol and the natural logarithm, respectively.
cGAN has become a popular choice for various image generation applications because of its exceptional performance. Pix2pix \citep{pix2pix_isola2017} is a supervised cGAN model that is specifically designed to condition an input image and generate a corresponding output image. In contrast to previous studies that have only worked for a specific application, pix2pix is a general-purpose image-to-image translation solution that consistently shows desirable performance in different tasks. For example, it can convert a black-and-white image into a color image or convert a satellite image into a map image. Pix2pix employs a ``U-Net" shaped architecture as generator and employs a convolutional ``PatchGAN" classifier as discriminator, which solely penalizes the structure at the scale of image patches. The combination of an adversarial loss and a loss $\ell_1$ during training allows pix2pix to generate high-quality output images that resemble their corresponding input images. Stimulated by its promising performance on diverse multimodal images, we adapt the generator and discriminator architectures from those of pix2pix. 

The solely deep learning-based SAR colorization work is proposed in \citep{diverse_sarcol_schmitt2018}. The network consists of two parts: a variational autoencoder (VAE) and a mixed density network (MDN), which aims to train a conditional color distribution from which different colorization hypotheses can be drawn. For the MDN training, they feed the features extracted from the \textit{conv7} layer of the pre-trained colorization network proposed by Zhang \citep{zhang2016colorful}, but not directly feeding the SAR image into the MDN. On one hand, the observed performance of this method is  not satisfying; on other hand, the procedure and implementation are very complicated compared with the general end-to-end CNN or GAN models. Hence, we decided to investigate an effective but simple end-to-end cGAN-based model inspired from pix2pix.

\section{Protocol} \label{protocol}
In this section, we propose a new protocol for the SAR colorization task. First, the protocol for colorizing SAR images is proposed allowing a supervised learning scheme and a reference-based performance assessment. Furthermore, three performance metrics are introduced to calculate the similarity between the output colorized images and the reference images.

\subsection{A protocol for colorizing SAR images}
Contrary to the image colorization task commonly found in computer vision community, SAR colorization lacks of  reference colorized SAR images to be used as ground-truth during training. To address this issue, we propose a new protocol for creating artificial color SAR images using the SAR-MS image fusion technique.

\begin{figure}[t]
	\centering
	\includegraphics[scale=0.35]{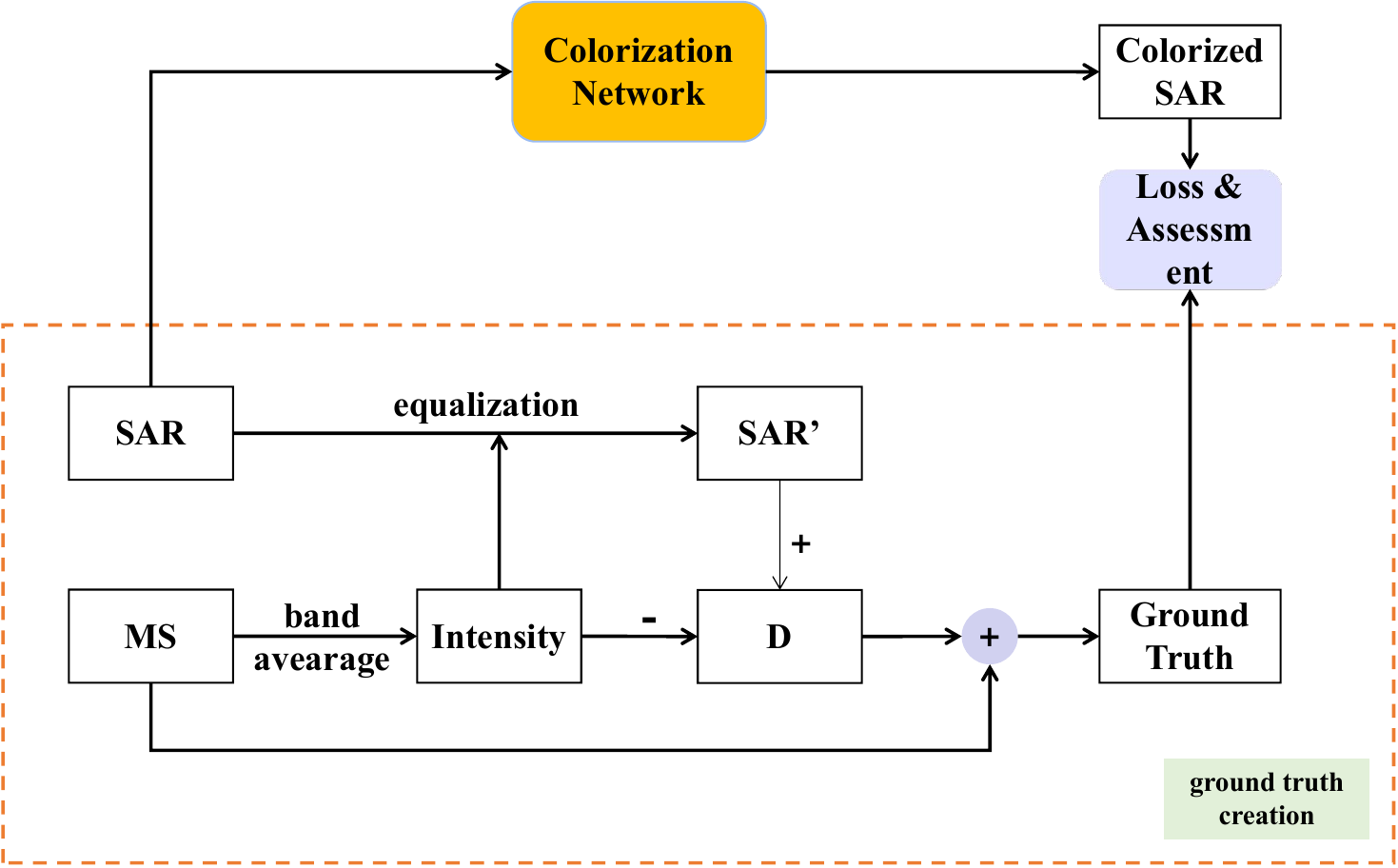}
	\caption{Outline of the protocol, the colorization and the assessment. } 
	\label{protocol_gt}
\end{figure}

\begin{figure}[t]
	\centering
	\includegraphics[scale=0.35]{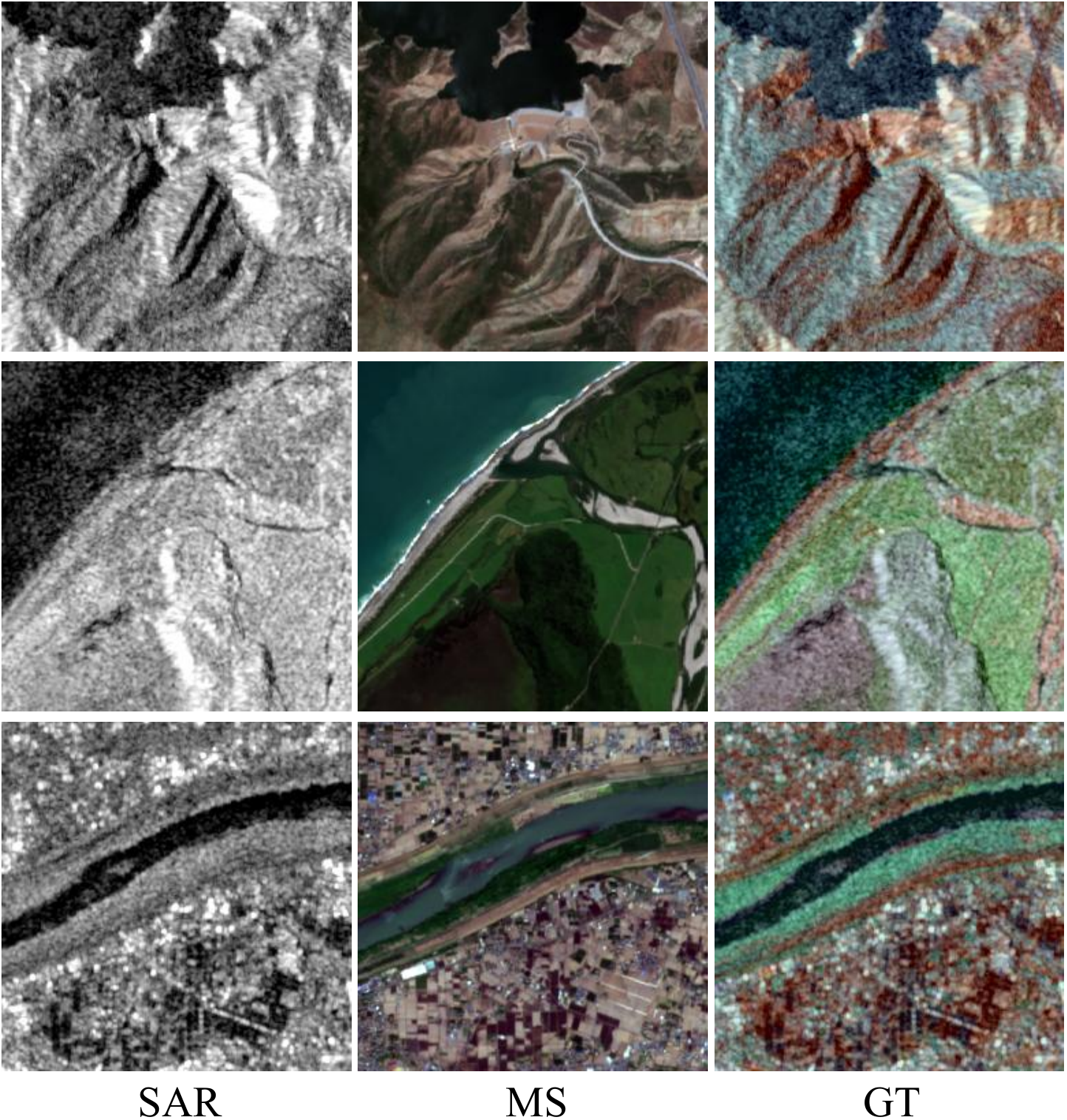}
	\caption{IHS fusion examples.} 
	\label{ihs_fusion_examples}
\end{figure}

Component substitution is a widely used class of fusion approaches \citep{pansharpeningreview_vivone2014}. One of the most representative methods in this family is intensity-hue-saturation (IHS), which relies upon the transformation of a red-green-blue (RGB) image into the IHS color space, where the spatial structure information is mainly contained in the intensity (I) component, while the spectral information is contained in the hue (H) and saturation (S) components. The intensity component is then replaced by the SAR image. Obviously, the higher the correlation between the SAR image and the intensity component, the lower the spectral distortion produced by the IHS method. Thus, before substitution, the SAR image should be histogram-matched to ensure comparable distributions between the latter and the intensity component. A powerful and lightweight approach, widely explored in the literature \citep{pansharpeningreview_vivone2020}, is based on linear matching by adjusting the mean and standard deviations of the distribution of the data to be substituted, i.e., the equalization of the first two moments of the distribution of the SAR image. The fusion process is then completed by applying an inverse transformation to bring the data back into the RGB color space. However, the IHS method is only suitable for processing RGB images, limiting its applicability for processing MS images. Tu \textit{et al.} \citep{gihs_tu2004} propose a generalization of the IHS methodology, called GIHS, which can be applied to images with more than three bands. GIHS is also referred to as fast IHS due to its computational efficiency, since it avoids sequential transformations, substitutions, and final backward step operations.
Assuming that the $\mathbf{MS}$ image is represented by three RGB spectral bands, we denote the intensity component of the IHS representation of $\mathbf{MS}$ as $\mathbf{I}$, and the average operation along the spectral dimension as $T$. The fast IHS-based SAR-MS fusion algorithm can be described as follows:
\begin{align}
	& \mathbf{I} = T(\mathbf{MS}), \\
	& \mathbf{SAR'} = (\mathbf{SAR} -  \mu_{\mathbf{SAR}}) \cdot \sigma_\mathbf{I} / {\sigma_{\mathbf{SAR}}} + \mu_\mathbf{I}, \\ 
	& \mathbf{D} = \mathbf{SAR'} - \mathbf{I},\\
	& \mathbf{GT} = \mathbf{MS} + \mathbf{D},
\end{align}
where $\mathbf{SAR}$ is the SAR image, $\mathbf{SAR'}$ is its histogram-matched version, $\mathbf{D}$ is the difference matrix between $\mathbf{SAR'}$ and the intensity component $\mathbf{I}$, $\mathbf{GT}$ is the result of image fusion (ground truth for our objective), $\mu_{\cdot}$ and $\sigma_{\cdot}$ are the average and standard deviation operators, respectively. Supervised training for SAR colorization can be easily implemented, and performance assessment can be carried out when ground truth is available. A graphical representation of the entire process can be found in Fig.~\ref{protocol_gt}. Several fusion results are illustrated in Fig.~\ref{ihs_fusion_examples}.

\subsection{Performance metrics}
The lack of a standardized quantitative evaluation scheme for SAR colorization is a well-known gap within the research community. In fact, the unique article based on supervised learning in the literature, \citep{diverse_sarcol_schmitt2018}, does not perform a numerical assessment. Instead, methods in similar tasks (e.g., SAR-to-optical image translation) borrow some common metrics used in the natural image processing community, such as structural similarity index (SSIM) \citep{ssim_wang2004image}, peak signal-to-noise ratio (PSNR), and mean squared error (MSE). In this work, we use of a set of remote sensing-based metrics to assess the performance of the problem at hand. Specifically, we propose the use of three metrics commonly employed in image fusion for remote sensing \citep{pansharpeningreview_vivone2021}: the normalized root mean square error (NRMSE) (that is the well-known erreur relative globale adimensionnelle de synth\'ese (ERGAS) index \citep{pansharpeningreview_vivone2021} but neglecting the resolution ratio between the images), the spectral angle mapper (SAM) \citep{sam_yuhas1992discrimination}, and the multidimensional (extended) version of the universal image quality index (Q4) \citep{uiqi_wang2002universal,q4_alparone2004global}. All these indexes take into account the spectral dimension of the colorized images, which is one of the most important aspects for colorization. Instead, metrics such as SSIM, PSNR, and MSE are indexes that work on single-band images. Even if they can be calculated for all the spectral bands of the colorized product and then averaged along the spectral dimension, spectral distortion (relevant for colorization) is not taken into consideration by these indexes with respect to the other ones. Thus, the suitability of our proposal is clear with respect to the previously adopted indexes and the problem at hand. 
More specifically, NRMSE is given by:
\begin{equation}
	NRMSE = \frac{1}{N} \sum_{n=1}^N\left(\frac{\operatorname{RMSE}(n)}{\mu(n)}\right),
\end{equation}
where RMSE represents the root mean square error between the ground-truth and the colorized image coming from a colorization approach, $\mu_n$ is the mean (average) of the $n$-th band of the ground-truth, and $N$ is the number of spectral bands (i.e. 3 for the problem at hand). Lower values of NRMSE indicate more similarity between the colorized SAR image and ground-truth data. The ideal value is 0.
SAM measures the spectral dissimilarity (spectral distortion) of the colorized image compared with the ground-truth. Assuming $\mathbf{x}$ and $\mathbf{y}$ are two spectral vectors, both having $C$ components, in which $\mathbf{x}=\left[x_1, x_2, \ldots, x_C\right]$ is the ground-truth spectral pixel vector and $\mathbf{y}=\left[y_1, y_2, \ldots, y_C\right]$ is the colorized SAR image spectral pixel vector, SAM is defined as the absolute value of the spectral angle between the two vectors in the same pixel, which can be given by:
\begin{equation}
	\operatorname{SAM}(\mathbf{x}, \mathbf{y})=\arccos \left(\frac{\langle\mathbf{x}, \mathbf{y}\rangle}{\|\mathbf{x}\|_2 \cdot\|\mathbf{y}\|_2}\right),
\end{equation}
where $\|\cdot\|_2$ denotes the norm $\ell_2$, $\arccos$ is the arccosine function, and $\langle \cdot, \cdot\rangle$ indicates the dot product. A global measurement of spectral dissimilarity is obtained by averaging the values in the image. The lower the value of the SAM index, the better the performance. The SAM value is usually measured in degrees and the ideal value is 0. 

Q4 is the four-band extension of the universal image quality index (UIQI), which has been introduced for the quality assessment of pansharpening \citep{q4_alparone2004global}. It can be calculated by:
\begin{equation}
	\mathrm{Q4}=\frac{4\left|\sigma_{\mathrm{z}_1 \mathrm{z}_2}\right| \cdot\left|\mu_{\mathrm{z}_1}\right| \cdot\left|\mu_{\mathrm{z}_2}\right|}{\left(\sigma_{\mathrm{z}_1}^2+\sigma_{\mathrm{z}_2}^2\right)\left(\mu_{\mathrm{z}_1}^2+\mu_{\mathrm{z}_2}^2\right)},
\end{equation}
where $\mathrm{z}_1$ and $\mathrm{z}_2$ are two quaternions, formed with spectral vectors of multi-band images, that is $\mathrm{z} = a + ib + jc + kd$, $\mu_{\mathrm{z}_1}$ and $\mu_{\mathrm{z}_2}$ are the means of $\mathrm{z}_1$ and $\mathrm{z}_2$, $\sigma_{\mathrm{z}_1 \mathrm{z}_2}$ denotes the covariance between $\mathrm{z}_1$ and $\mathrm{z}_2$, and $\sigma^{2}_{\mathrm{z}_1}$ and $\sigma^{2}_{\mathrm{z}_2}$ are the variances of $\mathrm{z}_1$ and $\mathrm{z}_2$. Because we only chose three bands from the Sentinel-2 multispectral images, we added an all-zero band (as the fourth band) to the ground truth and the colorized SAR image to fit the requirement of the Q4 index. The higher the value of the Q4 index, the better the quality. The optimal value is 1.

\subsection{Overview of colorization process}
Based on our protocol, colorization task training and testing can be easily performed. An overview of this process is given in Algorithm \ref{colorization_algorithm}.

\begin{algorithm}[H]
	\caption{Overview of colorization process.}
	\begin{algorithmic}
		\STATE 
		\STATE {\textsc{INPUT}} 
		\STATE \hspace{0.5cm} $\mathbf{X}:$ SAR image; $\mathbf{MS}:$ multispectral image. 
		
		\STATE {\textsc{OUTPUT}} 
		\STATE \hspace{0.5cm} $\hat{\mathbf{Y}}:$ Colorized SAR image. 
		
		\STATE {\textsc{PROTOCOL}} 
		\STATE \hspace{0.5cm} $\mathbf{Y} = IHS(\mathbf{X}, \mathbf{MS})$, where $IHS$ represents the classical IHS fusion technique and $\mathbf{Y}$ is used as ground truth in the next training step.
		
		\STATE {\textsc{TRAINING OF A BASELINE APPROACH}} 
		\STATE \hspace{0.5cm} $\mathbf{\Phi} = Training(\mathbf{Y}, \mathbf{X})$, where $\mathbf{\Phi}$ represents the training parameters of the model and $Training$ the specific training process depending on different baseline approach.
		
		\STATE 
		
		\STATE {\textsc{TESTING}} 
		\STATE \hspace{0.5cm} $\hat{\mathbf{Y}} = Colorization(\mathbf{X}, \mathbf{\Phi})$, where $Colorization$ represents the converged model after training.
		
		\STATE \hspace{0.5cm}\textbf{return}  $\hat{\mathbf{Y}}$
	\end{algorithmic}
	\label{colorization_algorithm}
\end{algorithm}

\section{Baseline and methodology} \label{baseline_method}
Considering the lack of established baseline methods, we have developed several baseline approaches for comparative analysis. These include spectral-based and spatial-spectral-based methods. Additionally, we have introduced a supervised conditional adversarial network method, which has demonstrated significant performance improvements. Moreover, the current state-of-the-art for SAR colorization using neural networks, i.e., DivColSAR \citep{diverse_sarcol_schmitt2018}, is introduced as baseline.

\begin{figure}[t]
	\centering
	\includegraphics[scale=0.3]{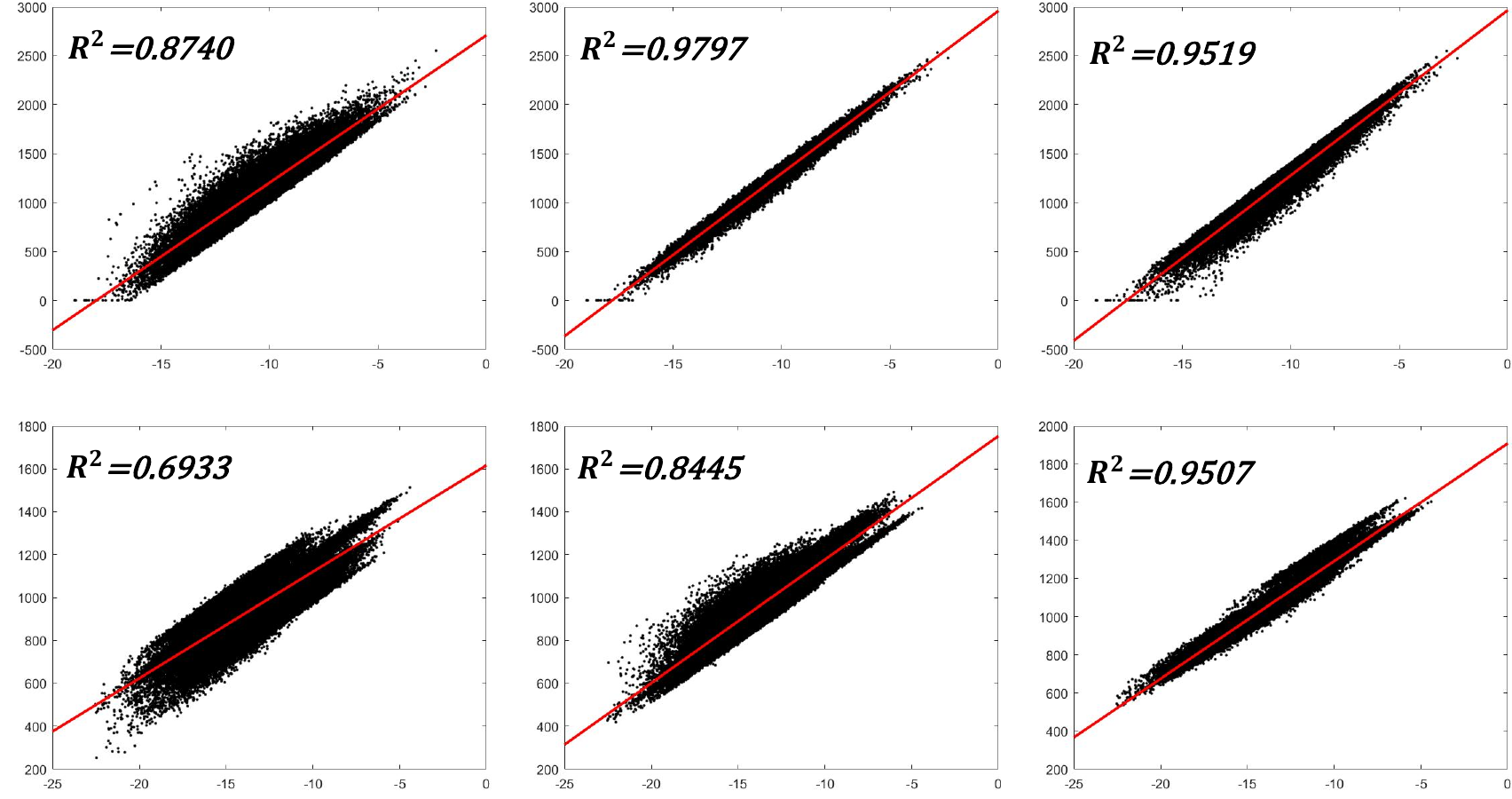}
	\caption{Data analysis with scatter plots. For each subfigure, $x$-axis and $y$-axis represent the value of the SAR band and one band of the corresponding ground-truth, respectively. $R^2$, ranging from 0 to 1, indicates the degree of goodness of fit between the variables, with higher values indicating stronger linear correlations.} 
	\label{data_analysis}
\end{figure}

\begin{figure}[t]
	\centering
	\includegraphics[scale=0.35]{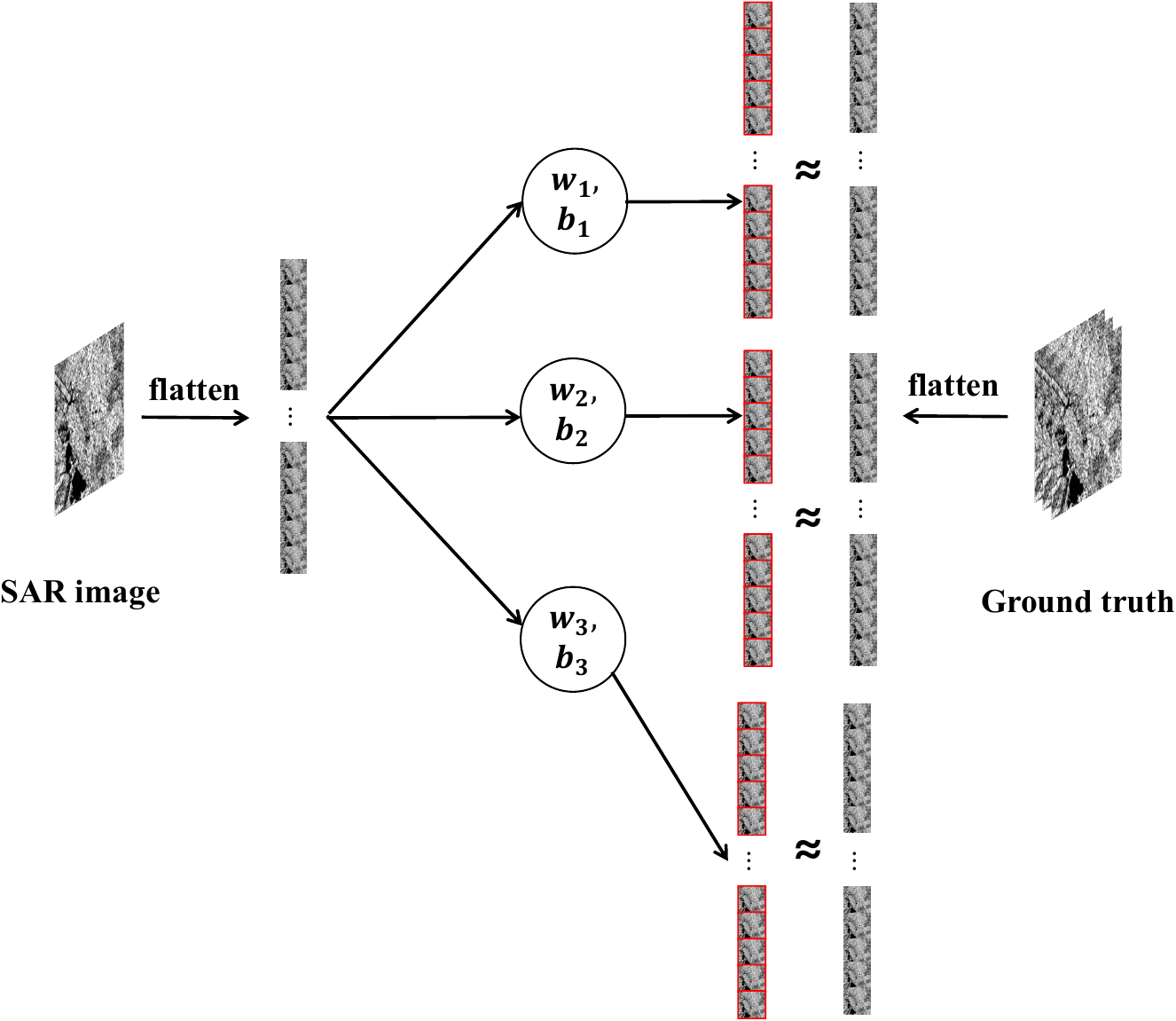}
	\caption{The overall architecture of LR4ColSAR.} 
	\label{lr4colsar}
\end{figure}

\begin{figure}[t]
	\centering
	\includegraphics[scale=0.4]{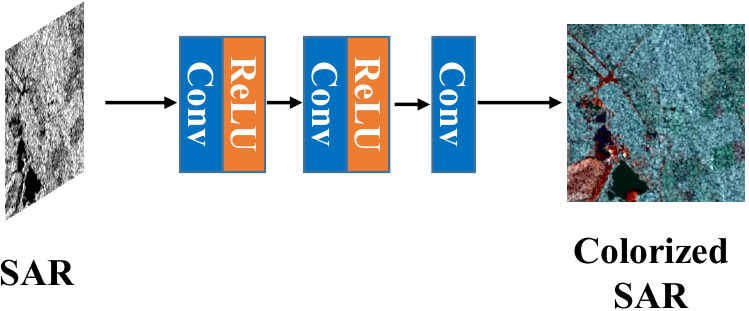}
	\caption{The overall architecture of CNN4ColSAR.} 
	\label{cnn4colsar}
\end{figure}

\begin{figure}[t]
	\centering
	\includegraphics[scale=0.25]{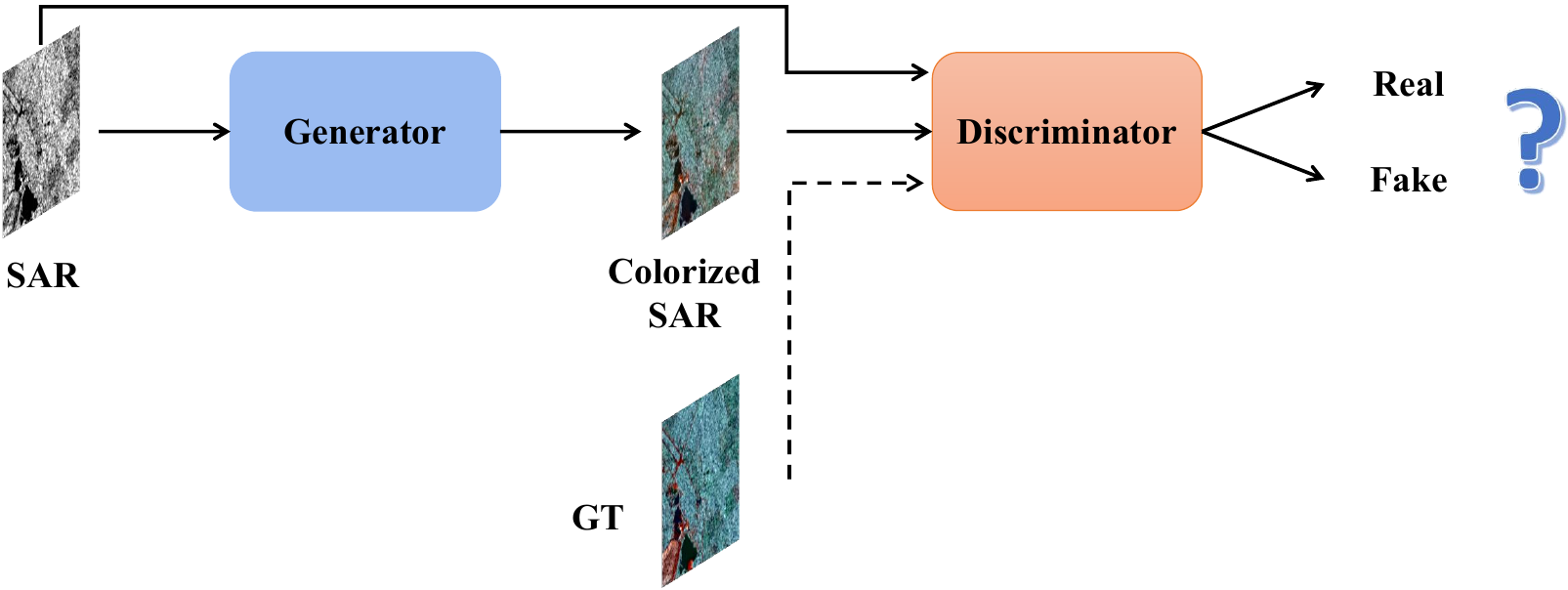}
	\caption{The overall architecture of cGAN4ColSAR.} 
	\label{overview_framework}
\end{figure}

\begin{figure*}[t]
	\centering
	\includegraphics[scale=0.35]{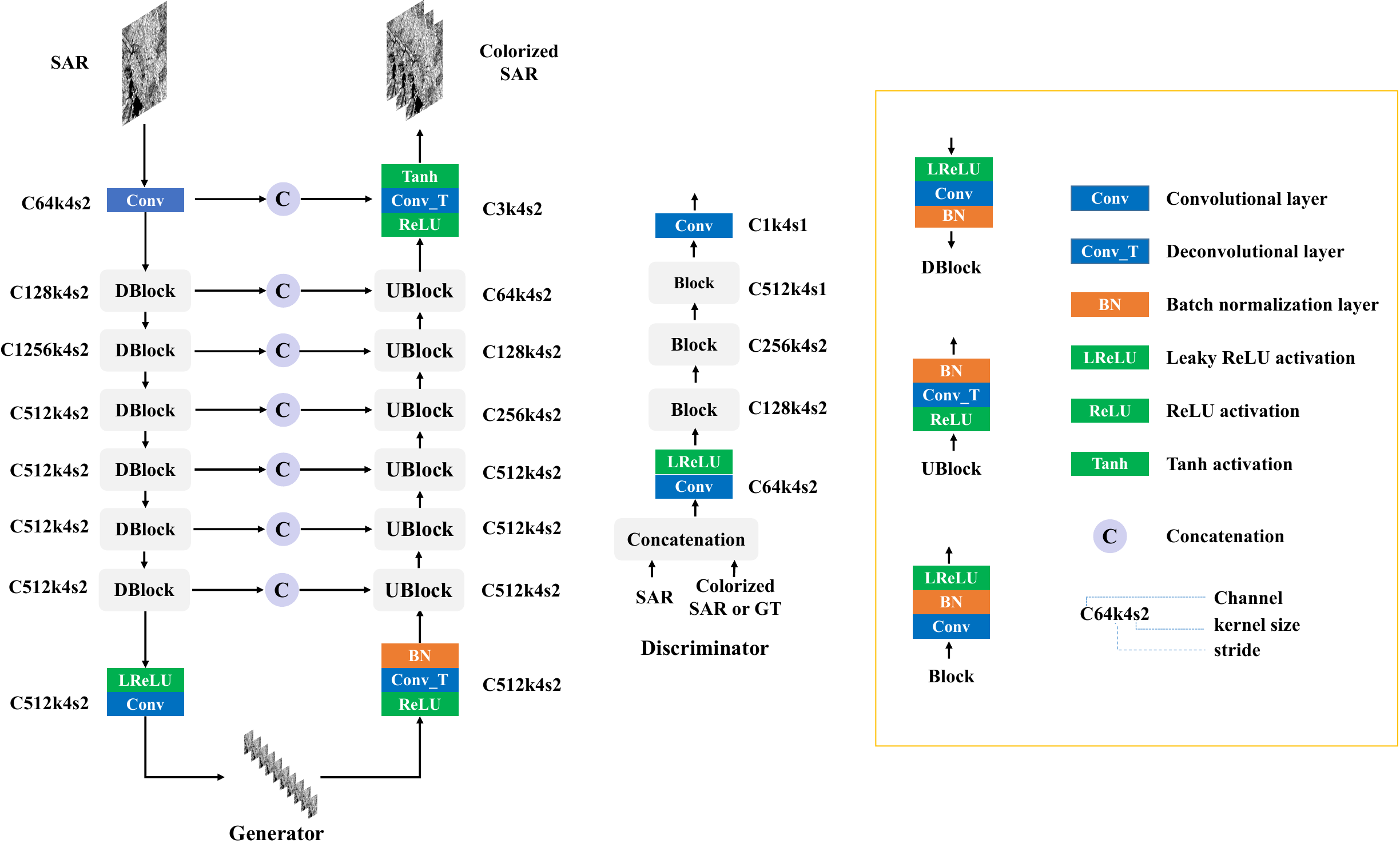}
	\caption{Details for cGAN4ColSAR.} 
	\label{detail_architecture}
\end{figure*}

\subsection{Spectral-based baselines for SAR colorization}
The objective of the SAR colorization task is to extract color information from paired MS images while retaining the spatial and radiometric information of SAR images. When a single-band image is considered as a unique entity, the task can be outlined as learning a linear or nonlinear mapping from the MS band image to the three-band MS image. The underlying assumption is that there exists a correlation/relationship between the SAR band and the MS bands. This kind of approaches represents the simplest solution for this problem. However, methods belonging to this family have several limitations when the relationship between the compared data is not clearly explainable through linear or non-linear models. Therefore, to validate this assumption, 2400 paired samples are selected for data analysis using scatter plots. Several examples of such analysis are presented in Fig.~\ref{data_analysis}, where the horizontal axis represents the values of the SAR band, and the vertical axis represents the values of the MS band. The regression line is in red. $R^2$ is a statistical measure that represents the variance proportion for a dependent variable that is explained by an independent variable in a regression model (in this case a linear regression model). Therefore, to be more practical, in this case, $R^2$ indicates how much the scatter plot follows a linear distribution (that is, given by linearly correlated data). Obviously, because optimal values are distributed along a line, the higher the $R^2$ value, the better the results. The optimal value is 1 obtained when the scatter plot is depicted as a line. The $R^2$ value of the corresponding example is marked in the upper left corner of each subfigure in Fig.~\ref{data_analysis}, and the mean and standard deviation of the $R^2$ values calculated on 2400 samples are 0.9193 and 0.0650, respectively. These results show that SAR and MS bands are approximately linearly correlated. Several (spectral-based) solutions mapping this correlation are proposed as follows.

\subsubsection{NoColSAR}
The simplest spectral-based solution relies on the replication of a single SAR band along the three channels. Although this approach does not capture any color, it does allow for preservation of spatial structure information. The NoColSAR method plays an important role as a baseline. By comparing this with the NoColSAR method, we are able to clearly observe how much the coloring effect of a colorization method has been improved (i.e., if the colorization can introduce proper colors).

\subsubsection{LR4ColSAR}
The second spectral-based solution is a linear regression method \citep{linear_regression_chatterjee1986, linear_regression_draper1998}, named LR4ColSAR, and depicted in Fig.~\ref{lr4colsar}. In this solution, the SAR band vector is represented by $\mathbf{x}$, while the predicted and target vectors are indicated as $\mathbf{\hat{y}}$ and $\mathbf{y}$, respectively. 
The adopted regression model is linear, and the coefficients are obtained through the minimum mean square error (MMSE) estimator.
\begin{align}
	& w^*_i,b^*_i = \arg\min_{w_i,b_i} \frac{1}{n} \sum_{j=1}^n [(w_{i}*x_j + b_{i}) - \mathbf{y}_j]^2, \label{train_lr4colsar} \\ 
	& \mathbf{\hat{y}}_i = w^*_i * \mathbf{x} + b^*_i, \label{test_lr4colsar}
\end{align}
where $i$ is the index of the multispectral band to be reconstructed starting from the SAR image, while $n$ corresponds to the length of the vectorized version of the SAR image and $j$ represents the $j$-th pixel. The weight and bias of the linear model are represented by $w_i$ and $b_i$, respectively. The result of this procedure is the set of biases and weights for each band, i.e. $w^*_i$ and $b^*_i$ for all $i \in [1,\dots,3]$, respectively.
To give a more clear presentation for the operation of this approach, we describe the process in Algorithm \ref{lr4colsar_algorithm}.

\begin{algorithm}[H]
	\caption{Colorization process of LR4ColSAR.} 
	\begin{algorithmic}
		\STATE 
		\STATE {\textsc{INPUT}} 
		\STATE \hspace{0.5cm} $\mathbf{X}:$ SAR image; $\mathbf{Y}:$ ground truth. 
		
		\STATE {\textsc{OUTPUT}} 
		\STATE \hspace{0.5cm} $\hat{\mathbf{Y}}:$ Colorized SAR image. 
		
		\STATE {\textsc{TRAINING}} 
		\STATE \hspace{0.5cm} Transform the SAR image and ground truth to vectors, which are denoted by $\mathbf{x}$ and $\mathbf{y}$, respectively. Then, training process is done with Equation (\ref{train_lr4colsar}).   
		
		\STATE 
		
		\STATE {\textsc{TESTING}} 
		\STATE \hspace{0.5cm} Testing process is done with Equation (\ref{test_lr4colsar}) to get the colorized SAR image vector and then transform the vector to image $\hat{\mathbf{Y}}$.
		
		\STATE \hspace{0.5cm}\textbf{return}  $\hat{\mathbf{Y}}$
	\end{algorithmic}
	\label{lr4colsar_algorithm}
\end{algorithm}

\subsubsection{NL4ColSAR}
In this section, we present a nonlinear regression technique as a further spectral-based solution, called NL4ColSAR, to capture the nonlinearity that the previous approach cannot map. This method is implemented using a simple neural network model. The neural network has one hidden layer and an output layer, with the number of neurons in each layer being 2 and 3, respectively. Additionally, the hidden layer is followed by a Tan-sigmoid type transfer function. The Levenberg-Marquardt algorithm is selected as a training algorithm and the mean square error ($\ell_2$ norm) is used as a loss function. The Tan-sigmoid transfer function can be formulated as follows.
\begin{equation}
f(x) = \frac{2}{1 + e^{-2x}} - 1,
\end{equation}
where $e$ is the exponential function.

The complete processing flow can be described as Algorithm \ref{nl4colsar_algorithm}.

\begin{algorithm}[H]
	\caption{Colorization process of NL4ColSAR.} 
	\begin{algorithmic}
		\STATE 
		\STATE {\textsc{INPUT}} 
		\STATE \hspace{0.5cm} $\mathbf{X}:$ SAR image; $\mathbf{Y}:$ ground truth. 
		
		\STATE {\textsc{OUTPUT}}
		\STATE \hspace{0.5cm} $\hat{\mathbf{Y}}:$ Colorized SAR image. 
		
		\STATE {\textsc{TRAINING}} 
		\STATE \hspace{0.5cm} Transform the SAR image and ground truth to vectors, which is denoted by $\mathbf{x}$ and $\mathbf{y}$, respectively. Then, train the neural network model by Levenberg-Marquardt algorithm and $\ell_2$ loss function.
				
		\STATE 
		
		\STATE {\textsc{TESTING}} 
		\STATE \hspace{0.5cm} Obtain the colorized SAR image vector and then transform the vector to image $\hat{\mathbf{Y}}$.
		
		\STATE \hspace{0.5cm}\textbf{return}  $\hat{\mathbf{Y}}$
	\end{algorithmic}
	\label{nl4colsar_algorithm}
\end{algorithm}

\subsection{Spatial-spectral based baseline}
The three spectral methods described above solve the colorization issue only from a spectral perspective, neglecting the spatial correlation between the SAR imagery and the ground-truth, thus often getting not satisfying results. To overcome this limitation, we introduce a convolutional neural network, which is well known for its powerful ability to capture local spatial features. Our proposed approach, named CNN4ColSAR, can be considered as a spatial-spectral solution. It consists of three convolutional layers, each of which is equipped with a ReLU activation function, with the exception of the final layer. CNN4ColSAR adopts a simple $\ell_1$ loss function. The CNN4ColSAR architecture is illustrated in Fig.~\ref{cnn4colsar} and the algorithm process can be expressed as Algorithm \ref{cnn4colsar_algorithm}.

\begin{algorithm}[H]
	\caption{Colorization process of CNN4ColSAR.} 
	\begin{algorithmic}
		\STATE 
		\STATE {\textsc{INPUT}} 
		\STATE \hspace{0.5cm} $\mathbf{X}:$ SAR image; $\mathbf{Y}:$ ground truth. 
		
		\STATE {\textsc{OUTPUT}}
		\STATE \hspace{0.5cm} $\hat{\mathbf{Y}}:$ Colorized SAR image. 
		
		\STATE {\textsc{TRAINING}} 
		\STATE \hspace{0.5cm} Update $\Phi$ with the Adam optimizer to minimize the $\ell_2$ loss function to learn the optimal solution.
		
		\STATE 
		
		\STATE {\textsc{TESTING}} 
		\STATE \hspace{0.5cm} Input the SAR image into the converged model to obtain the colorized SAR image $\hat{\mathbf{Y}}$.
		
		\STATE \hspace{0.5cm}\textbf{return}  $\hat{\mathbf{Y}}$
	\end{algorithmic}
	\label{cnn4colsar_algorithm}
\end{algorithm}

\subsection{DivColSAR}
We exploit the pioneering work, DivColSAR \citep{diverse_sarcol_schmitt2018}, as baseline. DivColSAR consists of a variational autoencoder (VAE) and a mixed density network (MDN). The VAE is trained first to generate a low level embedding of the ground truth. Based on the low-dimensional latent variable embedding, the MDN is then trained to generate a multimodal conditional distribution that models the relationship between the input SAR image and the embedding. By sampling from the generated distribution, the decoder of the VAE can generate a variety of colorized SAR images of the original gray-level SAR image.
We retained the original structure and parameter settings of this method, only making some minor adjustments where necessary. First, the ground truth is generated (coherently with our protocol) by the IHS fusion algorithm instead of the Lab transformation. Second, to compare this result with the other approaches in the benchmark, we applied the average of the top eight instances of the MDN-predicted Gaussian mixture distribution to have a unique colorized SAR image for the given SAR image in input. For fairness, the network has been retrained from scratch using the same training set as for the other approaches and the hyperparameters have been tuned, confirming the optimal configuration proposed in the original paper.

\subsection{Methodology: Image-to-image translation with conditional generative adversarial network for SAR colorization (cGAN4ColSAR)}
\subsubsection{Motivations of network design}
The image colorization task can be viewed as a scene recognition task followed by the assignment of specific colors to objects or regions, somewhat similar to image segmentation. However, preserving spatial and radiometric information while colorizing makes the colorization task more complex. Therefore, the network needs to focus on both global and local information to achieve satisfactory colorization results. 
U-Net \citep{unet_ronneberger2015} is a well-known and widely used network architecture in the field of image segmentation due to its exceptional performance. Its distinctive ``U" shape design is composed of a contracting path and a symmetric expanding path that enables precise localization and context capture, making it well-suited for the image colorization task. Specifically, in the SAR colorization task, the input SAR image and the output colorized SAR image have different surface appearances but share the same underlying structure. In particular, the location of prominent edges in the SAR image is roughly aligned with that in the colorized image. Therefore, it would be advantageous to incorporate this information using skip connections directly across the network.

\subsubsection{Architecture details of cGAN4ColSAR} 
In this study, we present a new conditional generative adversarial network, named cGAN4ColSAR, adapting the pix2pix generator and discriminator architectures. The architecture of the proposed model is illustrated in Figs.~\ref{overview_framework} and~\ref{detail_architecture}. The cGAN4ColSAR generator adopts a ``U"-shaped structure, which includes a contracting path and a symmetric expanding path. A skip connection is used between two symmetric layers to facilitate the transmission of low-level details. The input of the generator is the SAR image, and the output is the corresponding colorized SAR image. However, the discriminator is a full convolutional network composed of five layers. As a conditional GAN framework, it takes as input either the SAR image with the ground-truth or the output of the generator and is designed to distinguish between the true colorized SAR image and the fake one.

\textbf{Network architecture of generator.}
The generator in our proposed cGAN4ColSAR is made up of a contracting path and a symmetric expanding path, where each path consists of seven layers. The contracting path applies seven $4\times4$ filter size convolutions with stride 2 for downsampling. The number of convolution kernels is doubled at each downsampling step until it reaches a maximum of 512, while the first convolutional layer's feature channel number is 64. Each convolutional layer is followed by a leaky ReLU activation function and a batch normalization layer, except for the first and last layers, where the first layer is a single convolutional layer and the last layer lacks a batch normalization layer.
For the expansive path, the size of the feature map is doubled layer by layer by repeated use of $4\times4$ deconvolutions with stride 2. The number of feature channels is halved layer by layer, starting from the fifth layer, while the number of feature channels of the last convolutional layer is the same as the bands of the colorized SAR image. Each layer is followed by a ReLU activation function, a deconvolutional layer, and a batch normalization layer, except for the last layer, where the batch normalization layer is replaced by a Tanh activation function.

To preserve low-level image information, we add skip connections between the $i$-th layer of the contracting path and the $(9-i)$-th layer of the expansive path. Each skip connection concatenates all channels in the $i$ -th layer with those in the $(9-i)$ -th layer. Additionally, we do not include a dropout layer in our architecture, as its effect is limited according to the findings of pix2pix.

\textbf{Network architecture of discriminator.}
The discriminator proposed in our approach, based on the PatchGAN concept introduced in the pix2pix model, focuses on the local structure of the image patches to promote high-frequency detail capture in our approach. Our discriminator consists of five sequential convolution layers, each equipped with kernels of $4\times4$. The first three layers use a stride of 2 for downsampling, while the remaining layers use a stride of 1. The first convolution layer has 64 feature channels and is followed by a leaky ReLU activation function. The subsequent three layers have output feature maps of 128, 256, and 512, respectively, and each is followed by a batch normalization layer and a leaky ReLU activation function. The final layer is a single convolutional layer that produces a single feature map output.

\subsubsection{Loss function}
The purpose of this study is to solve the SAR colorization problem by presenting it as a conditional image generation task that can be effectively solved using conditional GAN. Specifically, we propose a generative network, denoted $G$ (using a set of parameters $\Theta_G$ to be trained), which is designed to map the distribution $p_{data}(\mathbf{X})$ to the target distribution $p_r(\mathbf{Y})$. By generating a colorized image $\hat{\mathbf{Y}}$ that is indistinguishable from the reference image $\mathbf{Y}$, evaluated by a discriminative network $D$ (using a set of parameters $\Theta_D$ to be trained) trained adversarially, our aim is to optimize a Min-Max objective function:
\begin{equation}
\begin{aligned}
	& \min _{\Theta_G} \max _{\Theta_D} \mathbb{E}_{\mathbf{X} \sim p_{\text {data }}(\mathbf{X}), \mathbf{Y} \sim p_r(\mathbf{Y})}\left[\log D_{\Theta_D}(\mathbf{X}, \mathbf{Y})\right] \\
	&+\mathbb{E}_{\mathbf{X} \sim p_{\text {data }}(\mathbf{X})}\left[\log \left(1-D_{\Theta_D}\left(\mathbf{X}, G_{\Theta_G}(\mathbf{X})\right)\right],\right.
\end{aligned}
\end{equation}
where $\log$ is the natural logarithm function. Using adversarial learning, a conditional GAN architecture can be used to produce accurate and realistic colorized SAR images. The generator network $G$ and the discriminator network $D$ are trained in an alternating manner. To optimize $G$, a combination of pixel loss and adversarial loss is used. To mitigate image blurring issues, the loss $\ell_1$ is utilized to calculate the absolute difference between the colorized SAR image and the ground truth as a pixel loss. The generator loss function, $\mathcal{L}(G)$, is made up of two components, the $\ell_1$ loss term and the adversarial loss term, which can be expressed as follows:
\begin{equation}
\begin{aligned}
	& \mathcal{L}(G) =\sum_{i=1}^B\left[-\log D_{\Theta_D}\left(\mathbf{X}, G_{\Theta_G}(\mathbf{X})\right)\right. \\
	&\left. + \alpha \left\|\mathbf{Y}-G_{\Theta_G}(\mathbf{X})\right\|_1\right], \label{loss_g}
\end{aligned}
\end{equation}
where $B$ represents the number of samples in a minibatch and $\alpha$ is a hyperparameter to weigh the contribution of the $\ell_1$ loss. In our experiments, the value of $\alpha$ is empirically set at 210.

The loss function, $\mathcal{L}(D)$, for the discriminator can be expressed as follows:
\begin{equation}
\mathcal{L}(D) = \beta \sum_{i=1}^B\left[\log(1 - D_{\Theta_D}\left(\mathbf{X}, G_{\Theta_G}(\mathbf{X})\right))+\log D_{\Theta_D}(\mathbf{X}, \mathbf{Y})\right], \label{loss_d}
\end{equation}
where the hyperparameter $\beta$ is used to weigh the contribution of the different loss terms. In our experiments, the value of $\beta$ is empirically set at 0.5.

The algorithm process is shown in Algorithm \ref{cgan4colsar_algorithm}.

\begin{algorithm}[H]
	\caption{Colorization process of cGAN4ColSAR.} 
	\begin{algorithmic}
		\STATE 
		\STATE {\textsc{INPUT}} 
		\STATE \hspace{0.5cm} $\mathbf{X}:$ SAR image; $\mathbf{Y}:$ ground truth. 
		
		\STATE {\textsc{OUTPUT}}
		\STATE \hspace{0.5cm} $\hat{\mathbf{Y}}:$ Colorized SAR image. 
		
		\STATE {\textsc{TRAINING}} 
		\STATE \hspace{0.5cm} Update $\Phi$ with the Adam optimizer to minimize loss function defined in Equation (\ref{loss_g}) and (\ref{loss_d}) to learn the optimal solution.
		
		\STATE 
		
		\STATE {\textsc{TESTING}} 
		\STATE \hspace{0.5cm} Input the SAR image into the converged model to obtain the colorized SAR image $\hat{\mathbf{Y}}$.
		
		\STATE \hspace{0.5cm}\textbf{return}  $\hat{\mathbf{Y}}$
	\end{algorithmic}
	\label{cgan4colsar_algorithm}
\end{algorithm}

\begin{figure*}[t]
	\centering
	\subfloat[]{\includegraphics[width=1.5in]{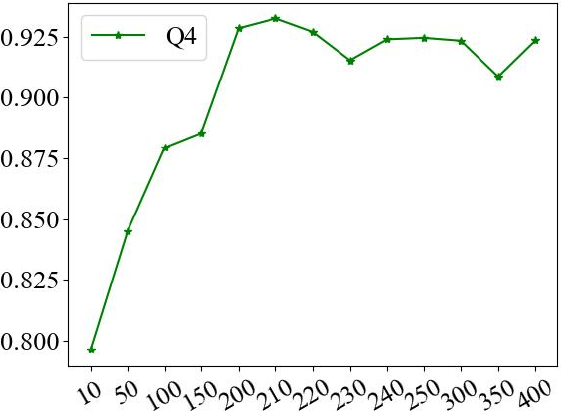}}
	\hfil
	\centering
	\subfloat[]{\includegraphics[width=1.5in]{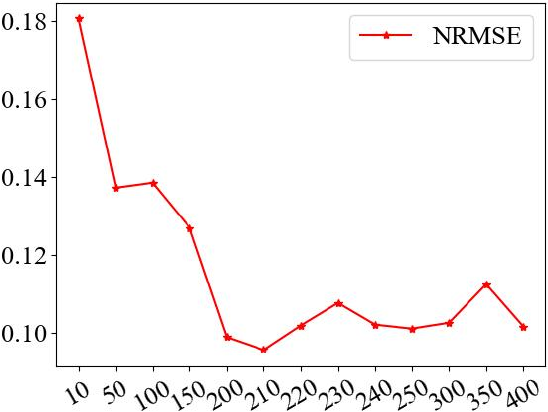}}
	\hfil
	\centering
	\subfloat[]{\includegraphics[width=1.5in]{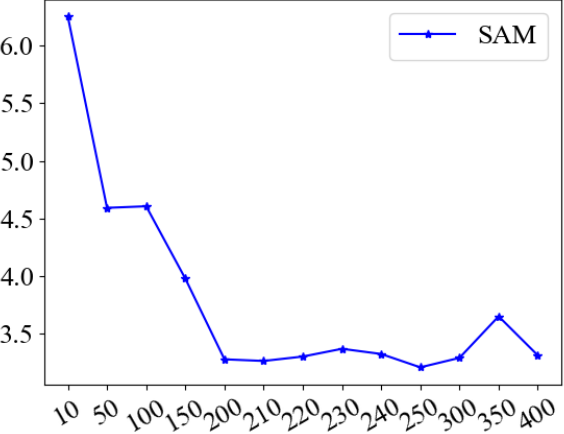}}
	\hfil
	\caption{Analysis by varying the parameter $\alpha$. The $x$-axis is represented by the values of $\alpha$. Figures from (a) to (c) show the results of Q4, NRMSE and SAM, repspectively.}
	\label{alpha_analysis}
\end{figure*}

\begin{figure}[t]
	\centering
	\includegraphics[scale=0.35]{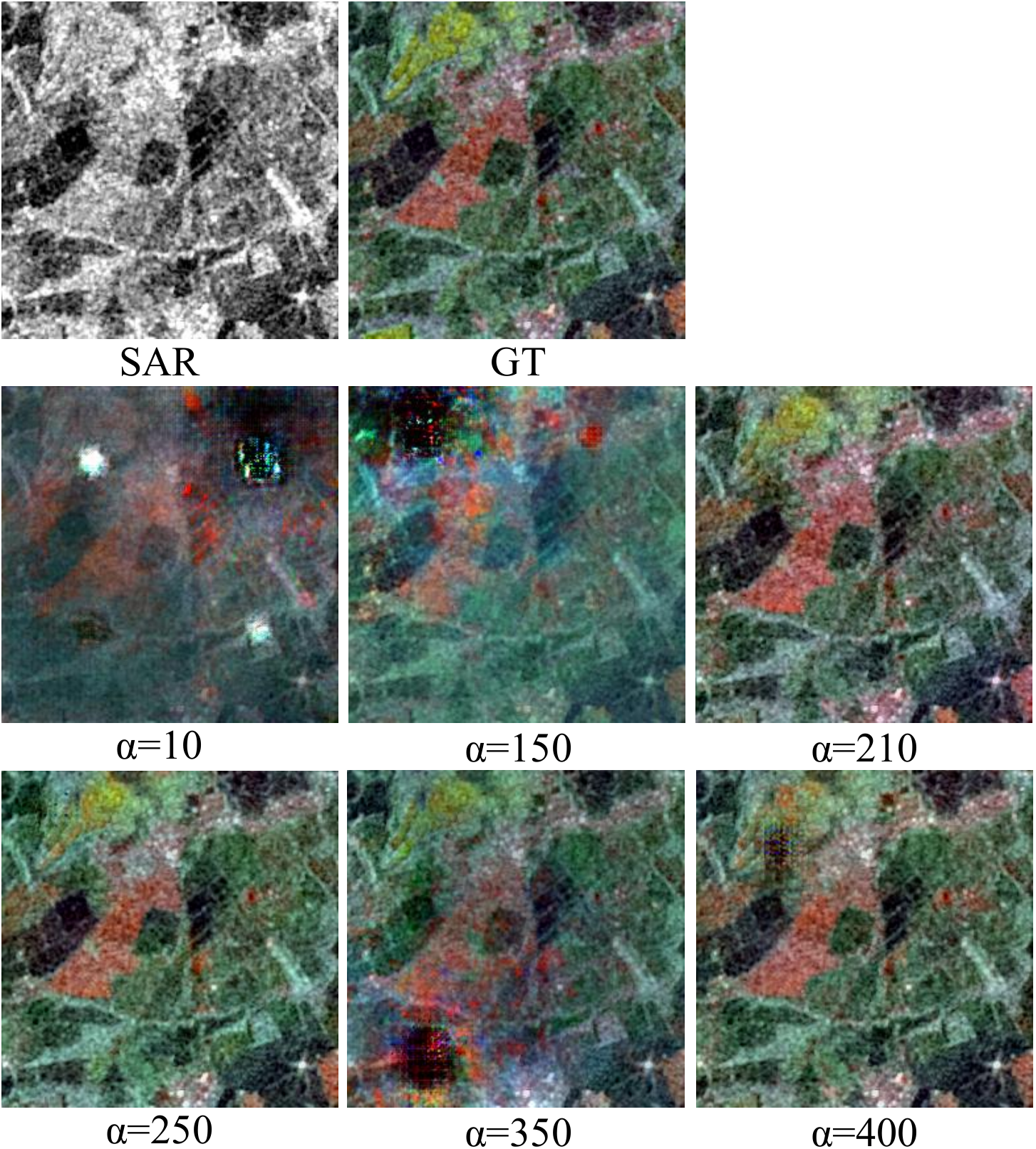}
	\caption{Visual results varying the parameter $\alpha$.} 
	\label{alpha_analysis_visual}
\end{figure}

\begin{figure}[t]
	\centering
	\includegraphics[scale=0.35]{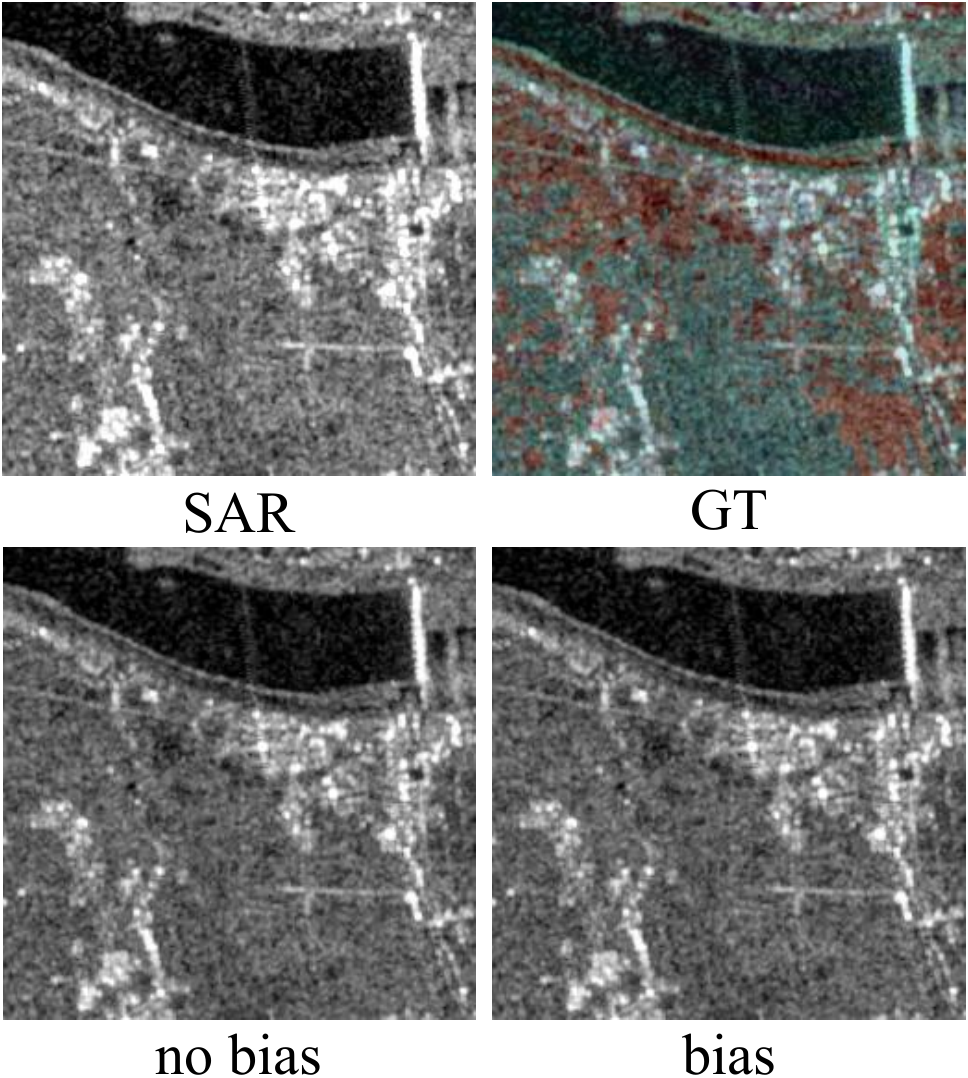}
	\caption{Model identification for LR4ColSAR. Visual results with and without the bias term.} 
	\label{wo_bias_visual}
\end{figure}

\begin{figure}[t]
	\centering
	\includegraphics[scale=0.35]{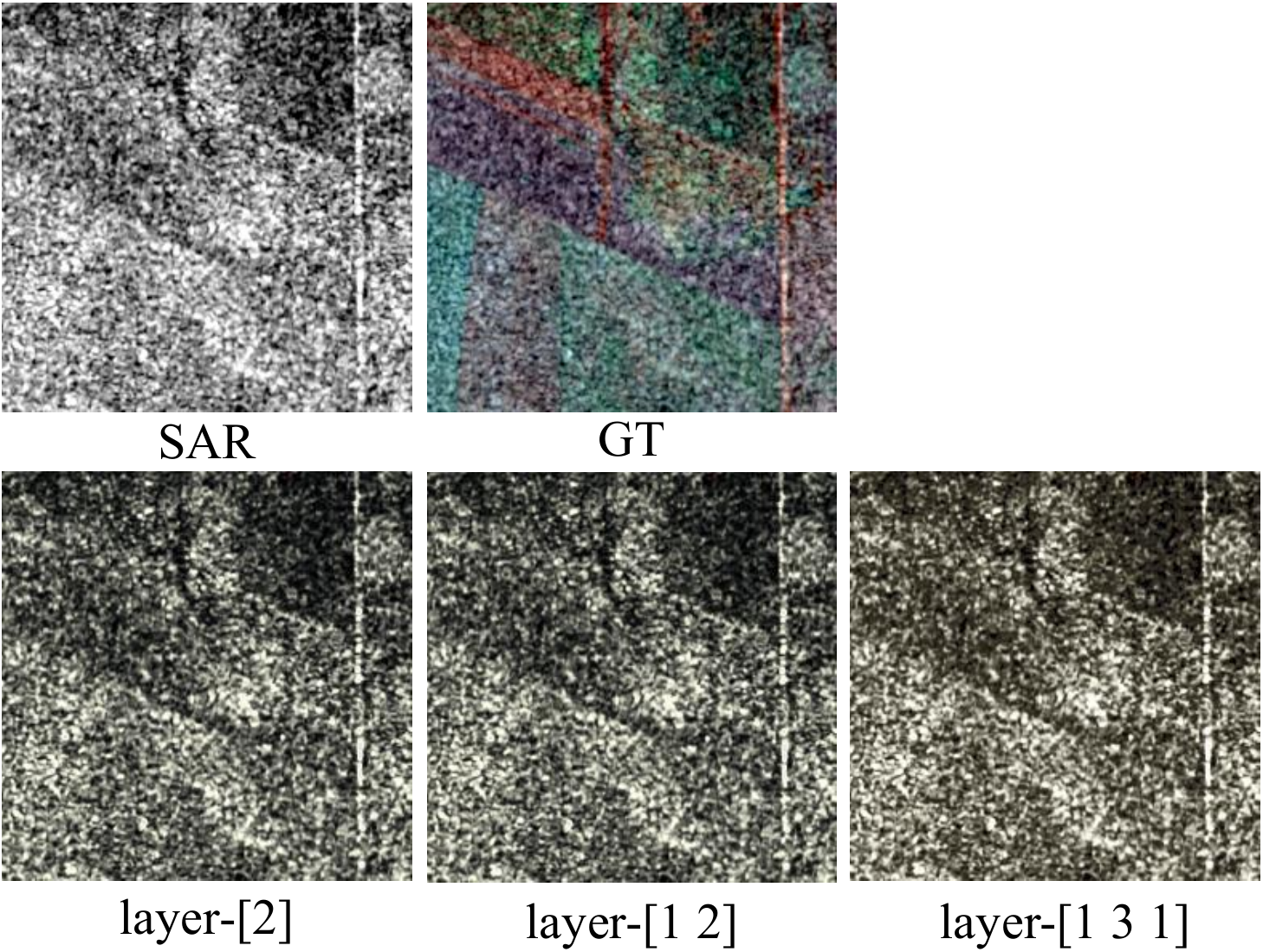}
	\caption{Visual results for different sets of parameters of NL4ColSAR. layer-[1 2] refers to a neural network architecture with two hidden layers, where the number of neurons in each layer is 1 and 2, respectively. Layer-[2] and Layer-[1 3 1] follow the same notation.} 
	\label{nl4colsar_variant_visual}
\end{figure}

\begin{figure*}[t]
	\centering
	\includegraphics[scale=0.35]{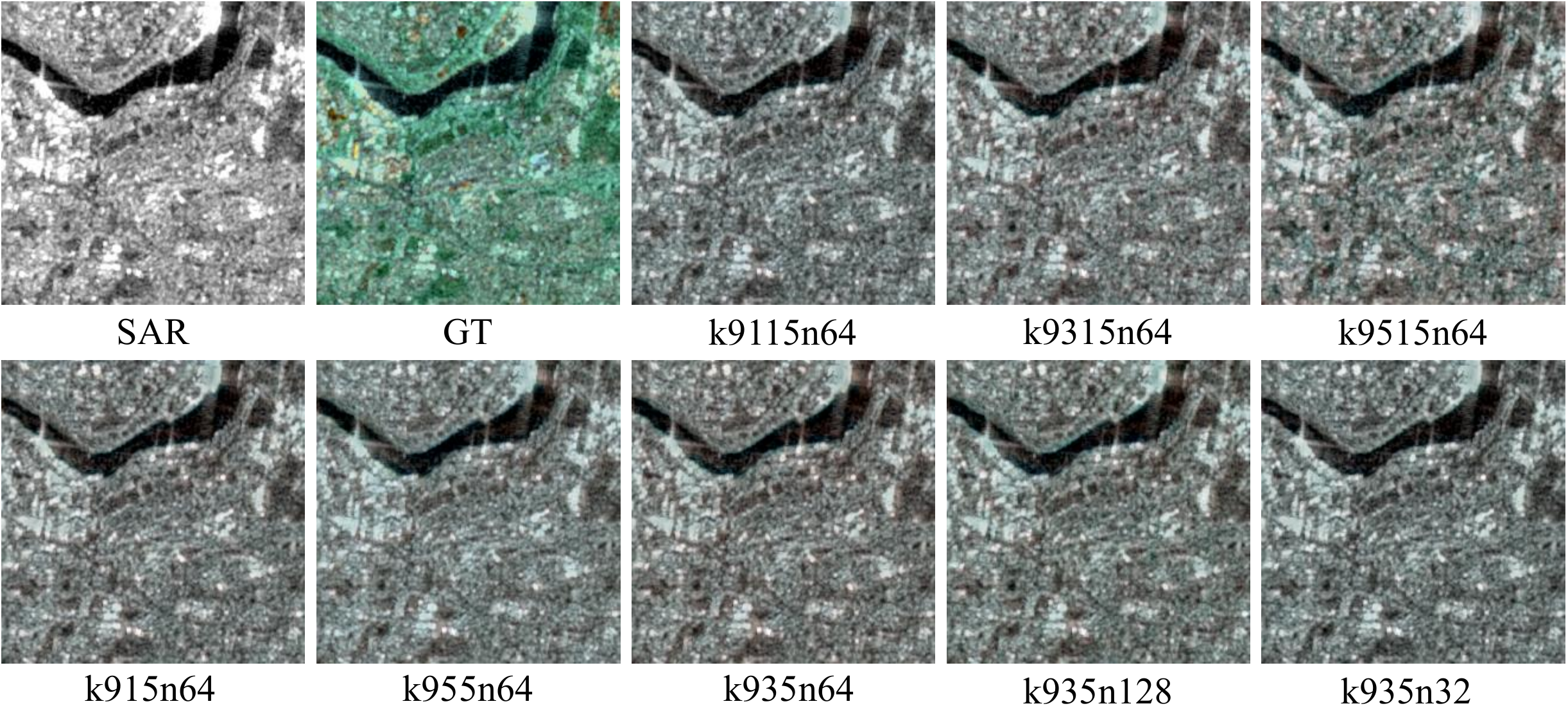}
	\caption{Visual results for different sets of parameters of CNN4ColSAR. k935n64 represents a convolutional neural network with three convolutional layers, where the kernel sizes of the convolutional layers are 9, 3, and 5, respectively, and the number of kernels in the first convolutional layer is 64. The same notation is used for the other names in the figure.} 
	\label{cnn4colsar_variant_visual}
\end{figure*}

\begin{figure}[t]
	\centering
	\includegraphics[scale=0.35]{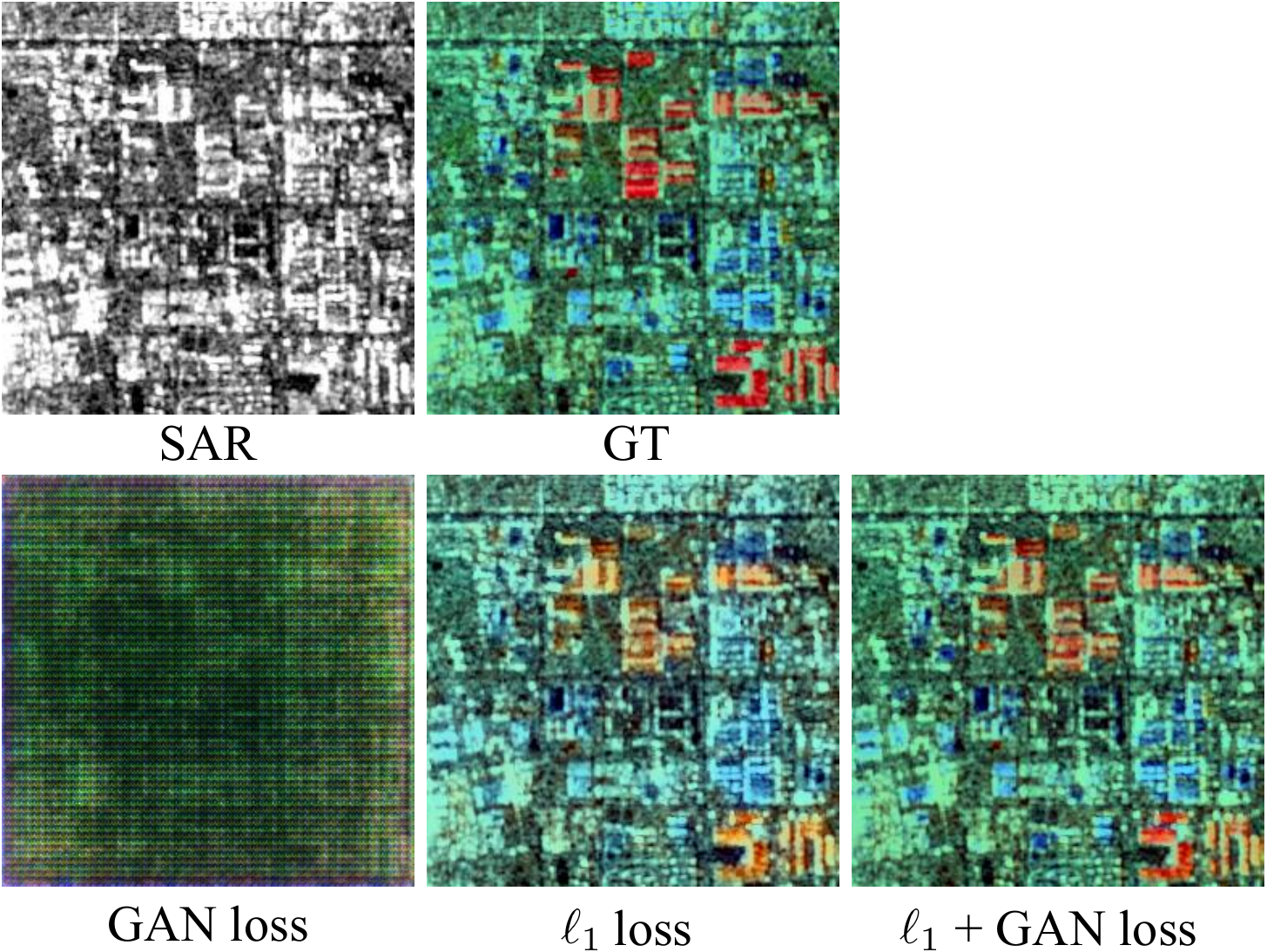}
	\caption{Visual results varying the composition of the loss function of cGAN4ColSAR.} 
	\label{cgan4colsar_loss_variant_visual}
\end{figure}

\begin{figure}[t]
	\centering
	\includegraphics[scale=0.35]{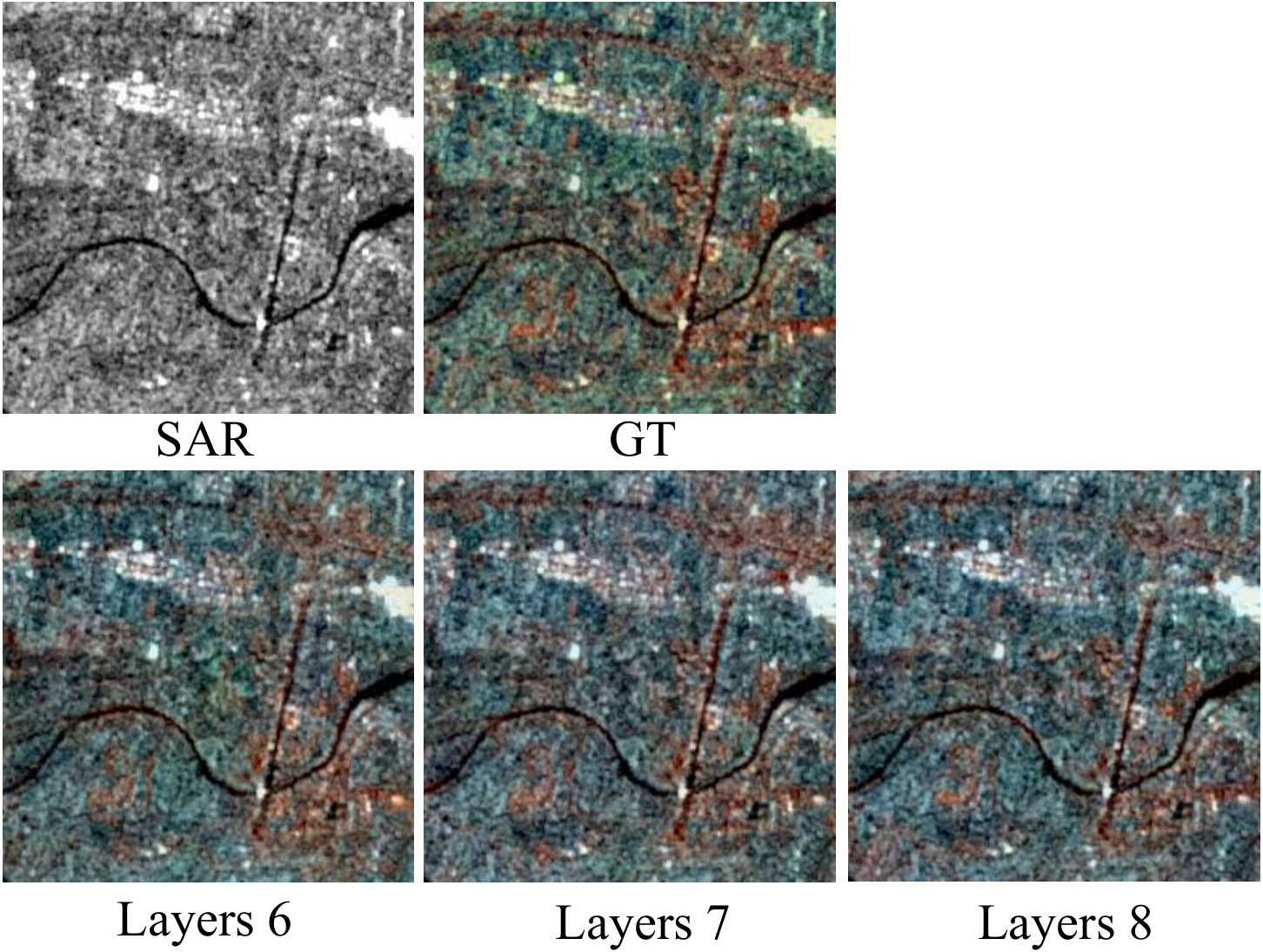}
	\caption{Visual results varying the number of layers of cGAN4ColSAR.} 
	\label{cgan4colsar_layer_variant_visual}
\end{figure}

\begin{figure*}[htpb]
	\centering
	\subfloat[]{\includegraphics[width=6in]{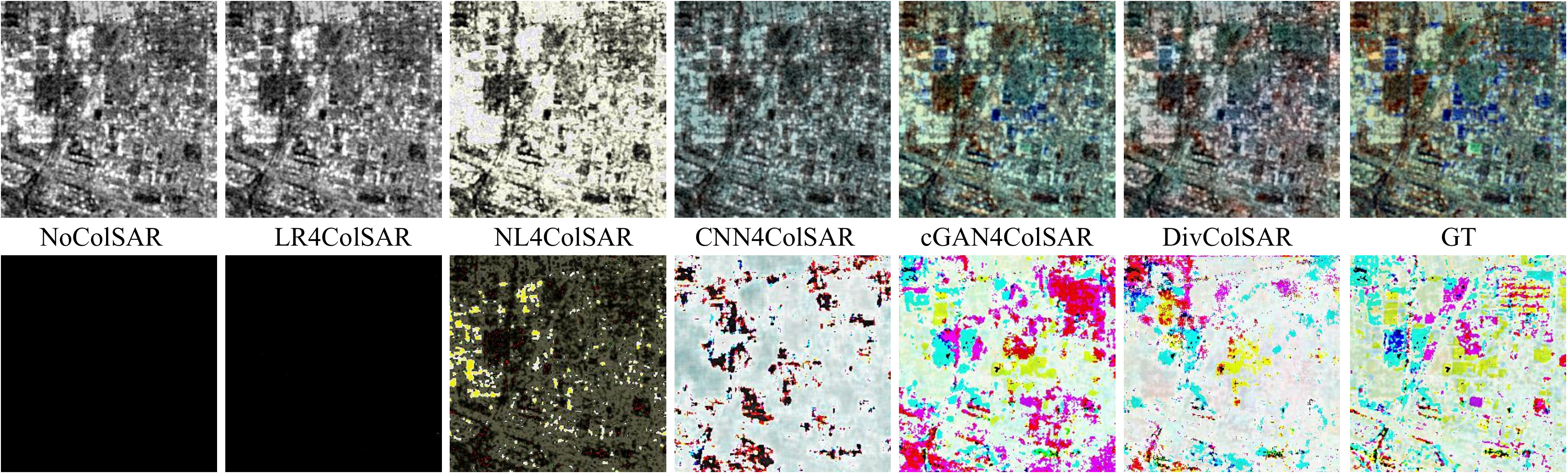}}
	\hfil
	\centering
	\subfloat[]{\includegraphics[width=6in]{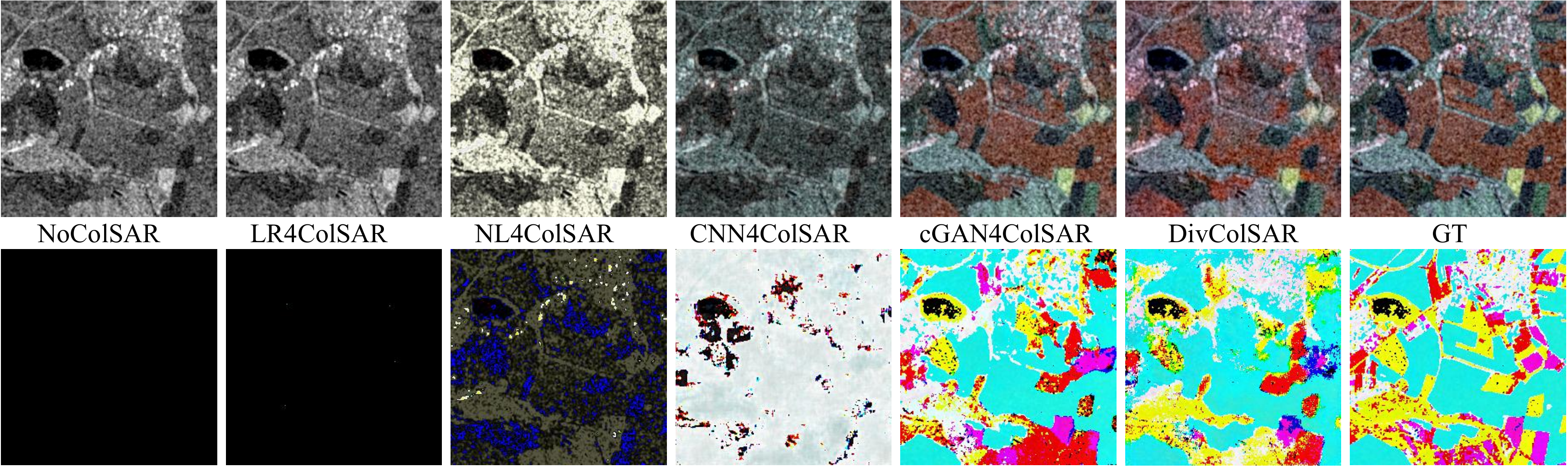}}
	\hfil
	\centering
	\subfloat[]{\includegraphics[width=6in]{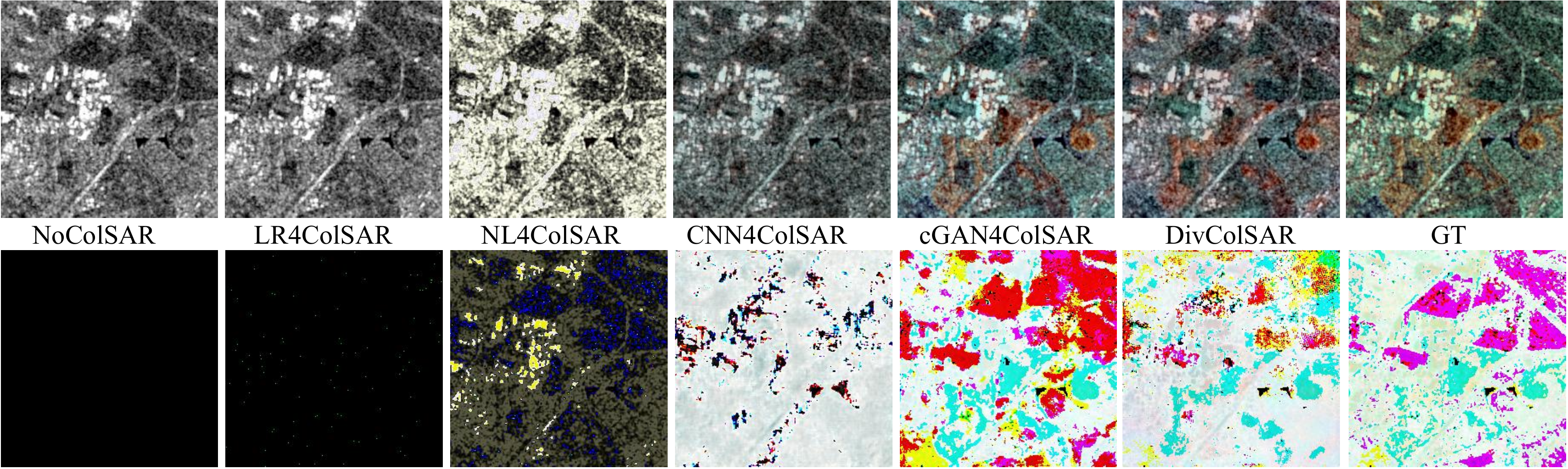}}
	\hfil
	\centering
	\subfloat[]{\includegraphics[width=6in]{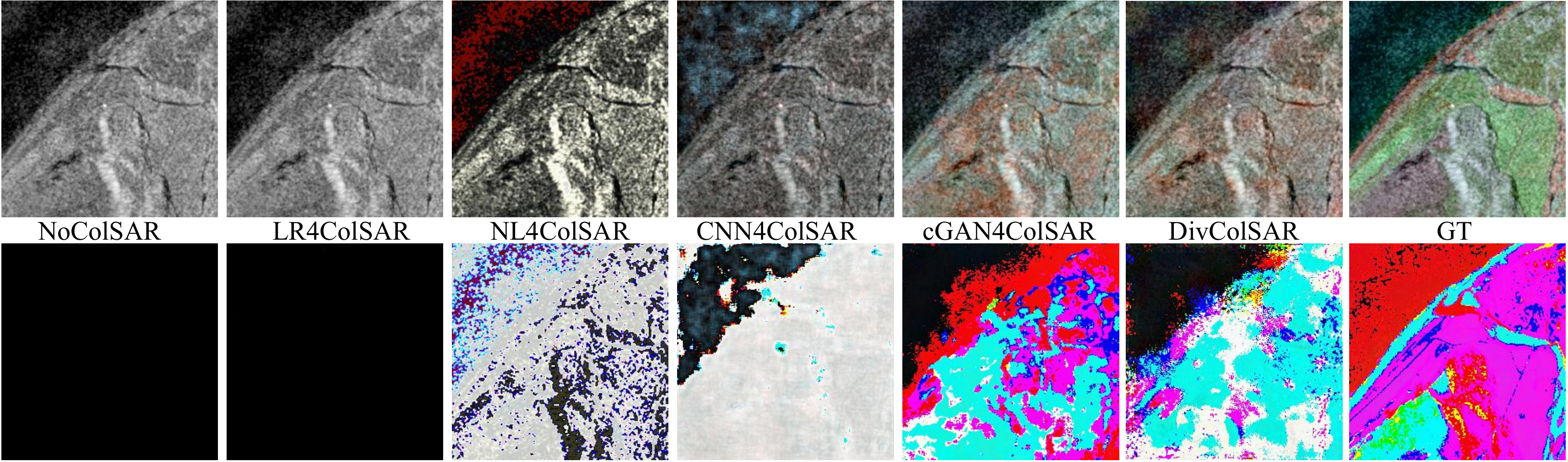}}	
	
	\caption{Figures from (a) to (d) show some examples of products obtained from the compared SAR colorization methods. Images in the second row in each subfigure depict the residuals between the colorized data (outcome of the colorization approach) and the corresponding NoColSAR image.}
	\label{comparison visual}
\end{figure*}

\begin{figure*}[t]
	\centering
	\includegraphics[scale=0.31]{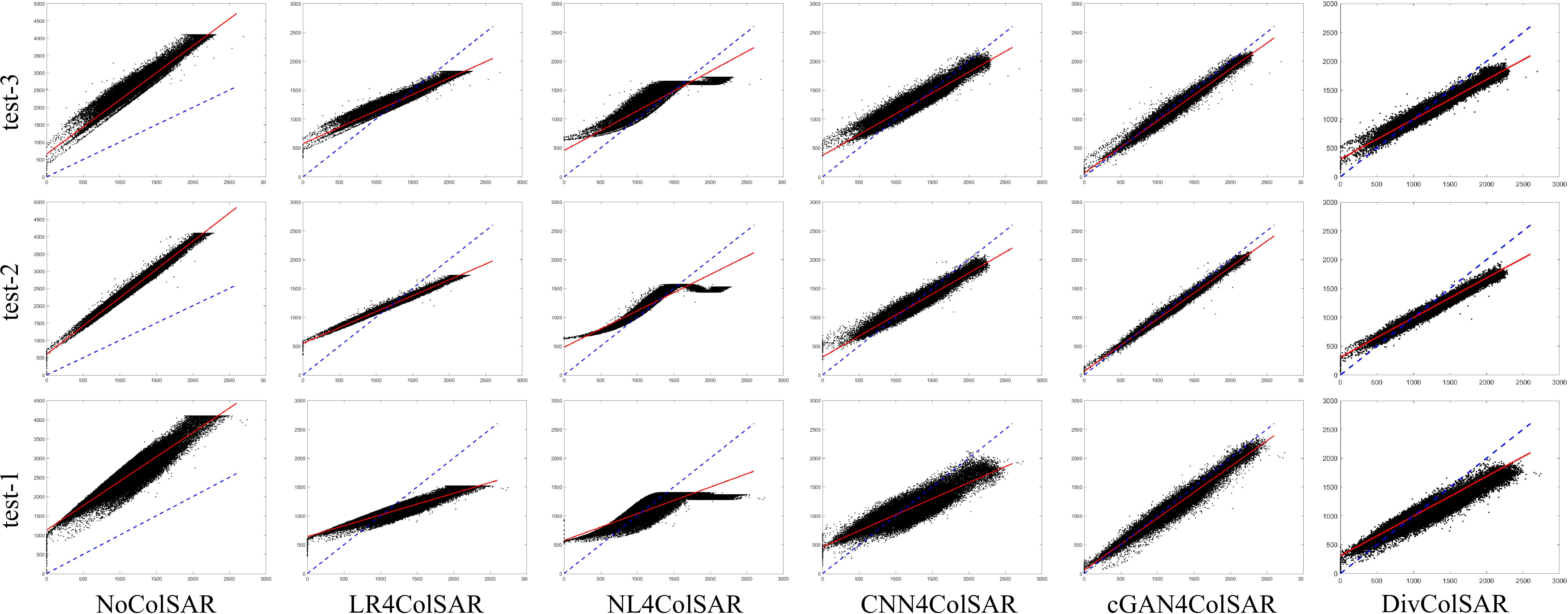}
	\caption{Scatter plots for the different SAR colorization methods on three test cases. $x$-axis and $y$-axis represent the values of ground-truth and the corresponding colorized SAR image, respectively. The red line and the blue dotted line represent the regression line and the optimal line (i.e., the quadrant bisector), respectively.} 
	\label{comparison_scatterplot}
\end{figure*}

\section{Results} \label{result}
This section will be dedicated to presenting the experimental analysis. The dataset will be presented first. Subsequently, the training details will be introduced. Finally, quantitative and qualitative results among the proposed methods and the pioneering method \citep{diverse_sarcol_schmitt2018} will be shown.

\subsection{Dataset} \label{dataset}
The dataset used in our study comes from SEN12MS-CR, a large-scale dataset for cloud removal provided by Ebel \textit{et al.} \citep{sen12mscr_ebel2021}. Each patch of SEN12MS-CR contains a triplet of orthorectified georeferenced Sentinel-2 images, including a cloudy and a cloud-free multispectral image, as well as the corresponding Sentinel-1 image. The multispectral Sentinel-2 image consists of thirteen bands, while the Sentinel-1 image comprises two polarimetric channels. The diversity of SEN12MS-CR is ensured, as it covers different topographies of the world over four seasons. For RGB optical images, we follow the standard practice of selecting the related bands from Sentinel-2 (i.e., 4, 3, and 2). Similarly, we use channel 1 (polarimetric VV) from the corresponding Sentinel-1 images to generate the gray-scale SAR image. Due to the large size of the SEN12MS-CR dataset, we randomly select a subset of 9663 SAR-MS image pairs for training and 800 pairs for testing. Basic information about this dataset is reported in Tab.~\ref{dataset_table}. It should be noted that the images in SEN12MS-CR remain in their raw bit depths, with 12 bit MS images and floating point SAR data. Before feeding the SAR images into the network, we perform a pre-processing step to adjust the SAR images in the range $[0,\dots, 2^p]$. Thus, we have the following:
\begin{equation}
\mathbf{SAR}_{adj} = \frac{\mathbf{SAR} - min(\mathbf{SAR})}{max\left(\mathbf{SAR} - min(\mathbf{SAR})\right)} \cdot 2^{p},
\end{equation}
where $\mathbf{SAR}$ and $\mathbf{SAR}_{adj}$ represent the SAR image and its adjusted version, respectively, $min$ and $max$ are the minimum and maximum operators, respectively, and $p$ is the bit depth of the Sentinel-2 image (i.e. 12).

\begin{table}[t]
	\centering
	\caption{Dataset information.}
	\resizebox{\linewidth}{!}
	{
		\begin{tabular}{ccc}
			\toprule
			Dataset & Training & Testing\\
			\midrule
			\multirow{3}{*}{SEN12MS-CR} & Patches: 9663 & Pathches: 800 \\
			& SAR size: 256×256 & SAR size: 256×256 \\
			& GT size: 256×256×3 & GT size: 256×256×3 \\
			\bottomrule
		\end{tabular}%
	}
	\label{dataset_table}%
\end{table}%

\subsection{Training details}
The three spectral-based methods, namely NoColSAR, LR4ColSAR, and NL4ColSAR, are implemented in Matlab R2015b. NoColSAR is a non-parametric method that requires no training, thus the colorization is directly conducted on the test set. LR4ColSAR and NL4ColSAR are based on regression models to retrieve the relationship between each colorized band and the SAR image. Specifically, LR4ColSAR adopts the least squares algorithm to estimate the parameters of the underlying linear regression model, while NL4ColSAR uses one hidden layer followed by a nonlinear transfer function (that is, the Tan sig) to map the nonlinear regression relationship. To train these regression models, 20 patches of size $256\times256$ are flattened to a vector, resulting in a sufficient amount of training data.
The two deep learning-based methods, CNN4ColSAR and cGAN4ColSAR, are implemented in Pytorch and are supported by a single NVIDIA Tesla V100 GPU. The batch size and learning rate are fixed at 8 and 1e-4, respectively, and the Adam optimizer is used to train the network from scratch. Training epochs of CNN4ColSAR and cGAN4ColSAR are both set to 300. CNN4ColSAR adopts a simple $\ell_1$ loss function. cGAN4ColSAR uses a loss function consisting of different loss terms, with the weight of the discriminator loss set at 0.5 and the weight of the loss set $\ell_1$ at 210. The compared method \citep{diverse_sarcol_schmitt2018} has been re-implemented and adapted for our data following the training details and setting the hyperparameters in agreement with the original paper.

\subsection{Quantitative and qualitative results comparison}

\begin{table}[t]\footnotesize
	\centering
	\caption{Numerical assessment for the proposed SAR colorization approaches. Best results are in boldface.}
	\begin{tabular}{rccc}
		\toprule
		Method & Q4   & NRMSE & SAM \\
		\midrule
		NoColSAR   & 0.3678±0.1563 & 1.0515±0.3905  & 6.4530±2.4532 \\
		LR4ColSAR  & 0.7400±0.1120 & 0.2294±0.0842  & 5.8586±1.8270 \\
		NL4ColSAR  & 0.7551±0.1000 & 0.2332±0.0874  & 5.8125±1.8048 \\
		CNN4ColSAR & 0.7929±0.0802 & 0.1994±0.0766  & 5.4443±1.8325 \\
		DivColSAR  & 0.839±0.091   & 0.2066±0.1274  & 6.0847±3.3086 \\
		cGAN4ColSAR & \textbf{0.9324±0.0655} & \textbf{0.0955±0.0643} & \textbf{3.2592±1.8803} \\
		\midrule
		ideal value & 1     & 0     & 0 \\
		\bottomrule
	\end{tabular}%
	\label{comparison}%
\end{table}%

\begin{table}[t]\footnotesize
	\centering
	\caption{The $R^2$ comparison for 2400 test cases. Best results are in boldface.}
	\resizebox{\linewidth}{!}{
		\begin{tabular}{ccccccc}
			\toprule
			Method & NoColSAR & LR4ColSAR & NL4ColSAR & CNN4ColSAR & cGAN4ColSAR & DivColSAR\\
			\midrule
			$R^2$ & 0.9193±0.065 & 0.9193±0.065 &  0.8492±0.076 & 0.8185±0.0868 &\textbf{0.957±0.0434} & 0.9291±0.0462\\
			\bottomrule
		\end{tabular}
	}
	\label{r_square}
\end{table}

In the previous sections, we introduced three spectral-based methods and two deep-learning-based methods. The quantitative and qualitative results of our proposed methods and the compared method DivColSAR \citep{diverse_sarcol_schmitt2018} are reported in Tab.~\ref{comparison} and depicted in Fig.~\ref{comparison visual}.
Although the NoColSAR method does not capture color information, it is used as a baseline to assess the improvements in colorizing provided by the other methods. Deep learning-based methods generally outperform spectral-based methods. cGAN4ColSAR shows the best performance, and it is much better than CNN4ColSAR and DivColSAR . This behavior can be attributed to the ``U" shape architecture, which captures both local and global information. CNN4ColSAR achieves the second best performance on NRMSE and SAM indexes while DivColSAR obtains the second best Q4 value. The performance of LR4ColSAR and NL4ColSAR is comparable. Instead, the baseline NoColSAR is the worst approach. 

For visual inspection, four test cases have been considered in Fig.~\ref{comparison visual}. To point out the improvements in the colorization task, we added the
residual images obtained by subtracting the colorized SAR image and the NoColSAR image. From Fig.~\ref{comparison visual}, the performance gap between the proposed cGAN4ColSAR and the rest of the benchmark is clear. DivColSAR can be considered the second-best as visual performance.

Finally, we analyze the results using scatter plots, as shown in Fig.~\ref{comparison_scatterplot}. More specifically, we flatten the colorized image and the corresponding ground-truth, and then report the scatter plots to check their similarity. To aid in visual inspection, the regression line is drawn in red, and the optimal line (i.e., the quadrant bisector) is in blue dotted. Optimal results lie in the optimal line. Thus, the best results have a scatter plot that is closer to the optimal line (and the related regression line should also lie on the optimal line). Moreover, in Tab.~\ref{r_square}, we show the mean and standard deviation of $R^2$ values for 2400 test cases. Analyzing both Fig.~\ref{comparison_scatterplot} and Tab.~\ref{r_square}, it is easy to see that cGAN4ColSAR clearly outperforms all other approaches for all three test cases. Instead, DivColSAR achieves the second-best performance.

\section{Discussion} \label{discussion}
In this section, we will carry out a series of experiments to analyze the impact of the hyperparameter settings and the different structures. More specifically, for LR4ColSAR, we will explore the influence of the bias coefficient; for NL4ColSAR, the network layer depth and the number of neurons per layer will be investigated; for CNN4ColSAR, the layer depth, the kernel size, and the numbers of filters will be analyzed; for cGAN4ColSAR, the influence of loss function and the network depth will be discussed.

\subsection{LR4ColSAR}
The LRColSAR model is a linear regression model that features three weighting and bias coefficients as learnable parameters. We exploit the model with bias coefficients because it exhibited superior performance in terms of all the metrics compared to the model without bias as shown in Tab.~\ref{wo_bias}.

\begin{table}[t]\footnotesize
	\centering
	\caption{LR4ColSAR performance with and without bias. Best results are in boldface.}
	\begin{tabular}{cccc}  
		\toprule
		\multicolumn{1}{r}{} & Q4   & NRMSE & SAM \\
		\midrule
		no bias  & 0.7329±0.1164 & 0.3546±0.1086  & 5.9646±1.8219 \\
		bias & \textbf{0.7400±0.1120} & \textbf{0.2294±0.0842 } & \textbf{5.8586±1.8270} \\
		\midrule
		ideal value & 1     & 0     & 0 \\
		\bottomrule
	\end{tabular}%
	\label{wo_bias}%
\end{table}%

Fig.~\ref{wo_bias_visual} illustrates the visual results of the LR4ColSAR model with and without bias coefficients. It can be seen that the performance of both variants is comparable, with no significant differences. Moreover, it seems that neither of the models can get useful color information.

\subsection{NL4ColSAR}
NL4ColSAR is based on multiple layers, each layer containing several neurons to map nonlinear relationships. A series of experiments is carried out to identify the most effective configuration of hidden layers and the corresponding number of neurons, as demonstrated in Tab.~\ref{nl4colsar_variant}. 
For the one hidden layer, the best performance is achieved when the number of neurons is 2. This choice gets the best values on all the three metrics. Taking a look at the solution with two hidden layers, setting 1-2 is the best. Instead, for the three hidden layers, the 1-3-1 setting exhibits the best performance. Among all configurations, the setting 2 is the best.
However, it can also be observed that the difference in the numerical results among the different settings is very small, which can be explained by the limited ability of representation. For each depth of the structure, we only chose the one with the best quantitative results to show it in Fig.~\ref{nl4colsar_variant_visual}. Visual results confirm the above statement.

\begin{table}[t]
	\centering
	\caption{Performance for the different architectures of NL4ColSAR. Best results are in boldface.}
	\resizebox{\linewidth}{!}
	{
		\begin{tabular}{ccccc}
			\toprule
			Hidden Layers & \multicolumn{1}{c}{Neurons} & Q4   & NRMSE & SAM \\
			\midrule
			\multicolumn{1}{c}{\multirow{10}[2]{*}{1 hidden layer}} & \multicolumn{1}{c}{1} & 0.7531±0.0999 & 0.2338±0.0872 & 5.8231±1.8283 \\
			& \multicolumn{1}{c}{2} & \textbf{0.7551±0.1000} & \textbf{0.2332±0.0874} & 5.8125±1.8048 \\
			& \multicolumn{1}{c}{3} & 0.7438±0.1049 & 0.2372±0.0880 & 5.8458±1.7848 \\
			& \multicolumn{1}{c}{4} & 0.7440±0.1028 & 0.2370±0.0878 & 5.8596±1.7735\\
			& \multicolumn{1}{c}{5} & 0.7445±0.1023 & 0.2362±0.0873 & 5.8506±1.7683 \\
			& \multicolumn{1}{c}{6} & 0.7459±0.1017 & 0.2360±0.0874 & 5.8662±1.7713 \\
			& \multicolumn{1}{c}{7} & 0.7397±0.1037 & 0.2382±0.0875 & 5.8700±1.7593 \\
			& \multicolumn{1}{c}{8} & 0.7395±0.1039 & 0.2382±0.0875 & 5.8714±1.7677 \\
			& \multicolumn{1}{c}{9} & 0.7395±0.1038 & 0.2383±0.0875 & 5.8802±1.7628 \\
			& \multicolumn{1}{c}{10}& 0.7394±0.1038 & 0.2383±0.0875 & 5.8809±1.7631 \\
			\midrule
			\multicolumn{1}{c}{\multirow{7}[2]{*}{2 hidden layers}} & 1-2   & 0.7465±0.1011 & 0.2354±0.0871 & 5.8579±1.8013 \\
			& 2-1   & 0.7415±0.1051 & 0.2374±0.0875 & 5.8234±1.8282 \\
			& 1-3   & 0.7442±0.1046 & 0.2371±0.0879 & 5.8448±1.7853 \\
			& 3-1   & 0.7424±0.1059 & 0.2371±0.0877 & 5.8120±1.8288 \\
			& 1-6   & 0.7454±0.1019 & 0.2363±0.0876 & 5.8648±1.7694 \\
			& 2-6   & 0.7390±0.1040 & 0.2384±0.0876 & 5.8820±1.7636 \\
			& 6-6   & 0.7380±0.1041 & 0.2386±0.0874 & 5.8814±1.7625 \\
			\midrule
			\multicolumn{1}{c}{\multirow{6}[2]{*}{3 hidden layers}} & 1-1-3 & 0.7445±0.1045 & 0.2370±0.0879& 5.8480±1.7838 \\
			& 1-3-1 & 0.7504±0.1022 & 0.2349±0.0877 & 5.8132±1.8285 \\
			& 3-1-1 & 0.7424±0.1058 & 0.2371±0.0878 & 5.8121±1.8287 \\
			& 1-1-6 & 0.7447±0.1020 & 0.2368±0.0877 & 5.8722±1.7667 \\
			& 1-6-1 & 0.7452±0.1026 & 0.2360±0.0876 & \textbf{5.8116±1.8288} \\
			& 6-1-1 & 0.7451±0.1024 & 0.2361±0.0875 & 5.8235±1.8281 \\
			\midrule
			\multicolumn{2}{c}{ideal value} & 1     & 0     & 0 \\
			\bottomrule
		\end{tabular}%
	}
	\label{nl4colsar_variant}%
\end{table}%

\subsection{CNN4ColSAR}
Parameters about kernel size, kernel numbers, and number of layers are investigated to find the best settings. Tab.~\ref{cnn4colsar_variant} reports the performance achieved as a function of the different combinations of parameters.
Let us assume that $K_i$ represents the kernel size of the $i$-th convolutional layer and $n_j$ represents the kernel numbers of the $j$-th convolutional layer. 
In the kernel size field on Tab.~\ref{cnn4colsar_variant}, for example, the value 9-1-5 means that the kernel size of the convolutional layer is 9, 1, 5 in a row, that is, $K_1=9$, $K_2=1$, and $K_3=5$. In the field filters in Tab.~\ref{cnn4colsar_variant}, for example, the 64-32-3 value represents the kernel numbers of each layer, that is, $n_1 = 64$, $n_2 = 32$, and $n_3 = 3$.

To analyze the sensitivity of the network to different kernel sizes, we set the kernel size of the second layer at (i) $K_2=1$, (ii) $K_2=3$, and (iii) $K_2=5$, while $K_1$ and $K_3$ are fixed to 9 and 5, respectively. The results in Tab.~\ref{cnn4colsar_variant} show that the use of a larger kernel size can improve the colorization performance. Specifically, the best Q4, NRMSE and SAM values are achieved by setting 9-5-5. The results suggest that expanding the receptive field is helpful, and thus the neighborhood information is used to improve the colorization performance.

Generally, a wider network will improve performance due to the increase of the representation capability. Thus, we also studied the impact of kernel numbers. We set three width levels, namely (i) $n_1=32$ and $n_2=16$; (ii) $n_1=64$ and $n_2=32$; (iii) $n_1=128$ and $n_2=64$. $c_3$ is fixed to 3 depending on the output image bands. The numerical results show that the a wider network can improve the colorization performance. Specifically, the widest level architecture obtains the best values for the NRMSE and SAM indexes, and the equally best value for the Q4 index.

In \citep{deeperbetter_he2015}, It is suggested that "the deeper the better," meaning that the network can benefit from moderately deepening the network. Based on the 9-1-5, 9-3-5, and 9-5-5 architectures with a middle-level kernel number, we added another convolutional layer with a kernel size of 1 and a kernel number of 32 before the last layer to deepen the structures. The deepened structures are 9-1-1-5, 9-3-1-5, and 9-5-1-5, as shown in Tab.~\ref{cnn4colsar_variant}. It is observed that deeper structures improve the performance, albeit slightly. 

As shown in Fig.~\ref{cnn4colsar_variant_visual}, despite the different architectures leading to distinct visual results, discerning the dissimilarity remains challenging. Specifically, for the colorization task, incorporating contextual information is crucial. However, in instances where the layers are shallow and the receptive field is limited, the network encounters difficulties in capturing global information. This observation may elucidate the reason for the relatively modest performance observed in various CNN4ColSAR variants. In order to strike a balance between performance and efficiency, we ultimately selected the k9515n64 structure as our optimal model. This structure comprises four layers with kernel sizes of 9-5-1-5 and the number of kernels for the first three layers set at 64, 32, and 32, respectively. 

\begin{table}[t]
	\centering
	\caption{Performance for different architectures of CNN4ColSAR. Best results are in boldface.}
	\resizebox{\linewidth}{!}
	{
    \begin{tabular}{ccccc}
	\toprule
	\multicolumn{1}{c}{Kernel size} & Filters & Q4   & NRMSE & SAM \\
	\midrule
	9-1-5 & \multicolumn{1}{c}{\multirow{3}[2]{*}{64-32-3}} & 0.7849±0.0822 & 0.2150±0.0753 & 5.6565±1.9902 \\
	9-3-5 &       & 0.7936±0.0800 & 0.2162±0.0743 & 5.6212±1.9263 \\
	9-5-5 &       & \textbf{0.7942±0.0807} & 0.2096±0.0762 & 5.6017±1.9424 \\
	\midrule
	\multirow{3}[1]{*}{9-3-5} & \multicolumn{1}{c}{32-16-3} & 0.7839±0.0788 & 0.2193±0.0750 & 5.6524±1.9704 \\
	\multicolumn{1}{c}{} & \multicolumn{1}{c}{64-32-3}  & 0.7936±0.0800 & 0.2162±0.0743 & 5.6212±1.9263\\
	\multicolumn{1}{c}{} & \multicolumn{1}{c}{128-64-3} & 0.7839±0.0788 & 0.2097±0.0730 & 5.5545±1.8864 \\
		\midrule
	9-1-1-5 & \multicolumn{1}{c}{\multirow{3}[1]{*}{64-32-32-3}} & 0.7867±0.0835 & 0.2135±0.0782 & 5.6557±1.9806 \\
	9-3-1-5 &       & 0.7854±0.0809 & 0.2100±0.0745 & 5.5980±1.8802 \\
	9-5-1-5 &       & 0.7929±0.0802 & \textbf{0.1994±0.0766} & \textbf{5.4443±1.8325} \\
	\midrule
	\multicolumn{2}{c}{ideal value} & 1     & 0     & 0 \\
	\bottomrule
	\end{tabular}%
	}
	\label{cnn4colsar_variant}%
\end{table}%

\subsection{cGAN4ColSAR}
There are two terms related to the loss function of cGAN4ColSAR, and we conduct experiments to analyze the performance sensitivity for the different combinations of the loss terms. The quantitative results are listed in Tab.~\ref{cgan4colsar_loss_variant}. It is clear that the use of the sole adversarial loss term is associated with a significant reduction of the performance. The $\ell_1$ loss term produces significantly better results than only the adversarial loss, indicating that the $\ell_1$ loss is a stronger constraint than the adversarial loss. Taking into account the comparable performance between the sole adversarial loss term and the two loss terms, the need of the GAN loss may be doubted. However, it is also observed that combining the two loss terms further improves the performance of all the indicators. Therefore, with the combination of $\ell_1$ and the adversarial losses, cGAN4ColSAR can achieve competitive results. Hence, the GAN loss is surely necessary. Qualitative results also verify the conclusion, as shown in Fig.~\ref{cgan4colsar_loss_variant_visual}. 

Considering the balance between effectiveness and efficiency, we explore architectures that vary the different levels of network depth. Specifically, shallow, middle, and deep levels are set to 6, 7 and 8, respectively. The adjustable layers are closed to the bottom layers, namely C512k4s2 DBlock and C512k4s2 UBlock. According to the numerical results shown in Tab.~\ref{cgan4colsar_layer_variant}, the performance becomes better as the layers increase and Layers 8 achieves the best performance. Visual inspection of the results are shown in Fig.~\ref{cgan4colsar_layer_variant_visual}. Thus, we selected Layers 8 as the optimal setting.

\begin{table}[t]
	\centering
	\caption{Variation of the loss function of cGAN4ColSAR. Best results are in boldface.}
	\resizebox{\linewidth}{!}
	{
	\begin{tabular}{ccccc}
		\toprule
		\multicolumn{1}{c}{$\ell_1$ loss} & GAN loss & Q4   & NRMSE & SAM \\
		\midrule
		 $\checkmark$     &  $\times$     & 0.9217±0.0669 & 0.1028±0.0634 & 3.4151±1.8242 \\
		\multicolumn{1}{c}{$\times$} & \multicolumn{1}{c}{$\checkmark$} & 0.5300±0.2034 & 0.4942±0.2954 & 10.8982±5.1114 \\
		 $\checkmark$     & \multicolumn{1}{c}{$\checkmark$} & \textbf{0.9324±0.0655} & \textbf{0.0955±0.0643} & \textbf{3.2592±1.8803} \\
		\midrule
		\multicolumn{2}{c}{ideal value} & 1     & 0     & 0 \\
		\bottomrule
	\end{tabular}%
	}
	\label{cgan4colsar_loss_variant}%
\end{table}%

\begin{table}[t]\footnotesize
	\centering
	\caption{Variation of the number of layers of cGAN4ColSAR. Best results are in boldface.}
	\begin{tabular}{cccc}
		\toprule
		Layers & Q4   & NRMSE & SAM \\
		\midrule
		6     & 0.9192±0.0661 & 0.1062±0.0648 & 3.3382±1.8163 \\
		7     & 0.9418±0.0624 & 0.0873±0.0618 & 3.2829±1.8528 \\
		8     & \textbf{0.9324±0.0655} & \textbf{0.0955±0.0643} & \textbf{3.2592±1.8803} \\
		\midrule
		ideal value & 1     & 0     & 0 \\
		\bottomrule
	\end{tabular}%
	\label{cgan4colsar_layer_variant}%
\end{table}%

Finally, we discuss the influence of the weight of the $\ell_1$ loss term of cGAN4ColSAR. Fig.~\ref{alpha_analysis} shows the results for each quality metric that fluctuates $\alpha$. It can be seen that the performance first increases and then slowly decreases as the value of the parameter $\alpha$ increases. These results are taken from visual inspection in Fig.~\ref{alpha_analysis_visual}. Thus, the best performance is obtained by $\alpha = 210$, which is used in our loss function.

\section{Conclusions and future developments} \label{conclusion}
In this work, we present a comprehensive research framework for SAR colorization based on supervised learning. This approach simplifies the utilization of supervised learning techniques and the evaluation of performance for the problem at hand. We have introduced three spectral-based methods, accompanied by textcolor{red}{three} deep learning-based techniques. The latter category includes a simple three-layer convolutional neural network, textcolor{red}{a model based on a variational autoencoder and mixture density network}, and an image-to-image model based on a conditional generative adversarial network. Furthermore, we propose a novel protocol for generating ground-truth samples for both training and testing deep learning-based approaches. As for quality metrics, we consider a set of well-known and widely used indices within the remote sensing image fusion community, namely Q4, SAM, and NRMSE. We conducted extensive experiments to explore the optimal hyperparameters for all the proposed methods. Subsequently, we conducted a thorough comparison among the proposed and optimized techniques from the benchmark to advocate for the cGAN-based solution for SAR colorization.

Future developments go towards the design of novel architectures devoted to the SAR colorization task based on generative models as the cGAN framework.

\section*{Acknowledgments}
This research was supported by the high performance computing (HPC) resources at Beihang University and the Supercomputing Platform of the School of Mathematical Sciences at Beihang University. This work was supported in part by the China Scholarship Council (CSC) under Grant 202206020138; in part by the National Natural Science Foundation of China under Grant 62371017 and in part by Academic Excellence Foundation of BUAA for PhD Students.

\bibliographystyle{elsarticle-harv} 
\bibliography{reference}


\end{document}